\documentclass[conference]{IEEEtran}
\usepackage{graphicx}

\usepackage[numbers]{natbib}
\usepackage{multicol}
\usepackage[colorlinks=true, allcolors=blue]{hyperref}
\usepackage{amsmath}
\usepackage{booktabs}
\usepackage{amsfonts}
\usepackage{ragged2e}
\usepackage{xspace}

\usepackage{makecell}
\usepackage[utf8]{inputenc}
\usepackage{wrapfig}
\usepackage{tabularx}
\usepackage{caption}
\usepackage{multirow}
\usepackage{pifont} 
\usepackage{subcaption}
\newcommand{\cmark}{\ding{51}}
\newcommand{\xmark}{\ding{55}}

\newcommand{\rewardname}{ratchet progress reward }

\begin{document}

\newcommand{\nickname}[0]{APEX\xspace}

\title{\nickname: Learning Adaptive High-Platform Traversal \\for Humanoid Robots}

\newcommand{\ProjectWeb}[0]{\href{https://apex-humanoid.github.io}{project webpage}}

\author{
Yikai Wang\textsuperscript{*1}, Tingxuan Leng\textsuperscript{*1}, Changyi Lin\textsuperscript{*1}, Shiqi Liu\textsuperscript{1}, \\ Shir Simon\textsuperscript{2}, 
Bingqing Chen\textsuperscript{2}, Jonathan Francis\textsuperscript{1,2}, Ding Zhao\textsuperscript{1}\\[0.1cm]
\textsuperscript{1}Carnegie Mellon University \quad  \textsuperscript{2}Bosch Center for Artificial Intelligence\\
}

\twocolumn[{
\renewcommand\twocolumn[1][]{#1}
\maketitle
\vspace{-0.6cm}
\begin{center}
\noindent\begin{minipage}{1.0\textwidth}
    \includegraphics[width= 1.0\linewidth]{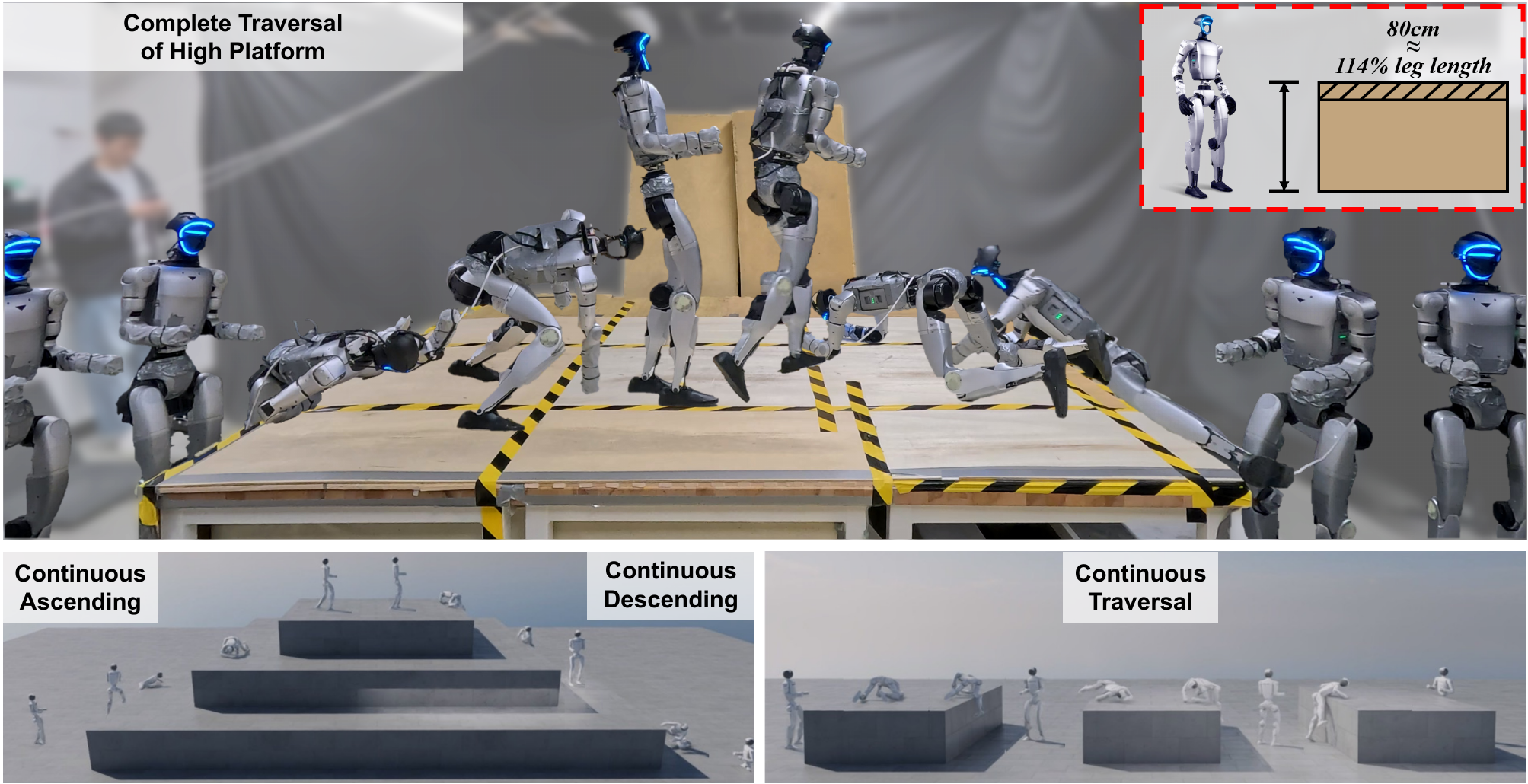}
    \end{minipage}\hspace{0.1in}
    \vspace{0.05in}
    
    \captionof{figure}{The robot adaptively traverses high platforms of up to $0.8\,\text{m}$ ($\approx 114\%$ of leg length) by leveraging diverse full-body behaviors, including climb-up, climb-down, stand-up, lie-down. Enabled by LiDAR-based elevation mapping, the policy exhibits context-aware whole-body coordination, allowing continuous and robust traversal across challenging terrain (\ProjectWeb).
}
    \label{fig:teaser}
    \vspace{0.0in}
\end{center}

}]

\begingroup
\renewcommand\thefootnote{}
\footnotetext[0]{*Equal contribution}
\addtocounter{footnote}{-1}
\endgroup

\begin{abstract}
Humanoid locomotion has advanced rapidly with deep reinforcement learning (DRL), enabling robust feet-based traversal over uneven terrain. Yet platforms beyond leg length remain largely out of reach because current RL training paradigms often converge to jumping-like solutions that are high-impact, torque-limited, and unsafe for real-world deployment.
To address this gap, we propose \emph{\nickname}, a system for perceptive, climbing-based high-platform traversal that composes terrain-conditioned behaviors: climb-up and climb-down at vertical edges, walking or crawling on the platform, and stand-up and lie-down for posture reconfiguration.
Central to our approach is a generalized ratchet progress reward for learning contact-rich, goal-reaching maneuvers. It tracks best-so-far task progress and penalizes non-improving steps, which provides dense yet velocity-free supervision, enabling efficient exploration under strong safety regularization.
Based on it, we train LiDAR-based full-body maneuver policies and reduce the sim-to-real perception gap via a dual strategy: training-time modeling of mapping artifacts and deployment-time filtering and inpainting of elevation maps.
Finally, we distill all six skills into a single policy that autonomously selects behaviors and transitions from local geometry and commands.
Experiments on a 29-DoF Unitree G1 humanoid demonstrate zero-shot sim-to-real traversal of $0.8\,\text{m}$ platforms ($\approx 114\%$ of leg length), with robust adaptation to platform height and initial pose and smooth, stable multi-skill transitions.
\end{abstract}

\IEEEpeerreviewmaketitle


\section{Introduction}
\label{sec:intro}

Locomotion is a fundamental capability for humanoid robots, yet has long remained challenging. Recent advances in deep reinforcement learning (DRL) have enabled robust, feet-based locomotion over uneven terrains~\cite{long2025learning, xue2025unified}. To further expand traversable terrain, prior systems have also learned whole-body jumping to get onto and off elevated structures~\cite{zhuang2024humanoid, zhang2024whole, ben2025gallant}.
However, jumping-based solutions typically achieve limited height (often below $63\%$ of leg length). Directly scaling them to substantially higher platforms (e.g., ledges or tables exceeding $100\%$ of leg length) requires large impulsive torques and induces high-impact dynamics, which can exceed actuator limits and pose unacceptable risk in real-world deployment.

For such extreme heights, a more reliable alternative is full-body climbing, which coordinates arms, torso, and legs to create distributed supports to ascend or descend in a controlled manner.
Building on climbing, complete high-platform traversal involves multiple behaviors: climb-up and climb-down at vertical edges, walking or crawling on the platform, stand-up and lie-down for posture reconfiguration between prone and upright configurations.
Despite its promise, learning and executing high-platform traversal presents two key challenges.

First, the four full-body maneuvers (climb-up, climb-down, stand-up, and lie-down) are difficult to learn with DRL.
In contrast to cyclic, command-conditioned locomotion (e.g., walking, running, or crawling), where tracking objectives such as base velocity and periodic contacts provide dense supervision, these maneuvers are contact-rich and goal-reaching. Success is defined by satisfying terminal conditions through staged contact transitions and whole-body reconfiguration (e.g., moving the lower body and center of mass above the platform during climb-up). Their contact patterns and motion velocities vary across phases and depend on perceived geometry, making such dense locomotion-style tracking rewards ill-defined.

Second, complete high-platform traversal is a long-horizon sequential problem that requires autonomous skill selection and seamless switching. A unified end-to-end policy must (i) acquire a diverse repertoire of skills, (ii) infer the appropriate behavior from local terrain observations and high-level user commands, and (iii) trigger smooth transitions at the correct moments while maintaining robot stability. These requirements couple perception, high-level decision-making, and low-level control across extended, multi-contact interaction phases, substantially increasing the difficulty of learning.

To address these challenges, we propose \emph{\nickname}, a system for learning \textit{adaptive high-platform traversal} based on a two-stage framework.
First, we train a library of six skills via DRL: the four goal-reaching full-body maneuvers and two cyclic locomotion skills.
To make the goal-reaching maneuvers learnable and deployable, we introduce a generalized \emph{ratchet progress} reward that maintains a self-updating best-so-far task state and penalizes the agent unless it strictly surpasses this state.
This yields dense, task-aligned supervision while remaining velocity-free, enabling efficient exploration under strong safety regularization and preventing ``retrace'' exploitation.
For perception, we leverage a LiDAR-based elevation mapping pipeline and bridge the sim-to-real gap with a dual strategy: training-time modeling of mapping artifacts and deployment-time filtering and inpainting of raw maps.
To make the learned skills ready for seamless transitions, we improve distribution matching between the predecessor's terminal states and the successor's initial states with strategies on reward design and data sampling.
We then distill the six teacher policies into a unified student policy, training on a mixture of skill-focused and transition-focused environments.
With these design choices, our unified policy achieves complete traversal of a $0.8\,\text{m}$ platform ($\approx 114\%$ of leg length) with zero-shot sim-to-real transfer on a 29-DoF Unitree G1 humanoid robot, and remains robust under variations in platform height and initial robot pose.
We further conduct extensive comparisons showing that the proposed ratchet progress reward is critical for learning adaptive contact-rich maneuvers.

In summary, our contributions are:
\begin{itemize}
\item A two-stage learning framework for adaptive high-platform traversal that integrates contact-rich full-body maneuvers and cyclic locomotion into a single controller.
\item A generalized ratchet progress reward that provides dense, velocity-free supervision for learning deployable contact-rich, goal-reaching maneuvers.
\item The first humanoid traversal policy that achieves real-world traversal over platforms exceeding $114\%$ of leg length, demonstrating robust adaptation, autonomous skill selection, and smooth transitions.
\end{itemize}


\section{Related Work}
We review prior works in three aspects most relevant to our system, and summarize key distinctions in Tab.~\ref{tab:related_comparison}.

\begin{table}[ht]
\setlength{\tabcolsep}{2.6pt}
\centering
\caption{Comparison with existing methods on humanoid full-body maneuver capabilities. Extreme height is defined as platforms exceeding $100\%$ of leg length.}
\begin{tabular}{c|c|c|c|c|c}\toprule
\multirow{3}{*}{Methods} & \multirow{3}{*}{\begin{tabular}[c]{@{}c@{}} Parkour \\ \cite{zhuang2024humanoid, zhang2024whole, ben2025gallant} \end{tabular}} & \multirow{3}{*}{\begin{tabular}[c]{@{}c@{}} Stand-Up \\ \cite{chen2025hifar, he2025learning, huang2025learning} \end{tabular}} & \multirow{3}{*}{\begin{tabular}[c]{@{}c@{}} Trajectory \\ Tracking \\ \cite{yang2025omniretarget} \end{tabular}} & \multirow{3}{*}{\begin{tabular}[c]{@{}c@{}} Motion \\ Generation \\ \cite{xu2025parc} \end{tabular}} & \multirow{3}{*}{\begin{tabular}[c]{@{}c@{}} \textbf{\nickname} \\ \textbf{(Ours)} \end{tabular}} \\ 
&&&&& \\
&&&&& \\
\midrule
Extreme height & \xmark & \xmark & \cmark & \cmark & \cmark \\
Full-body contact & \xmark & \cmark & \cmark & \xmark & \cmark \\
Terrain perception & \cmark & \xmark & \xmark & \cmark & \cmark \\
Terrain/pose-adaptive & \cmark & \cmark & \xmark & \cmark & \cmark \\
Reference-free & \cmark & \cmark & \xmark & \xmark & \cmark \\
Real robot & \cmark & \cmark & \cmark & \xmark & \cmark \\
Unified multi-skill & \cmark & \xmark & \xmark & \xmark & \cmark \\
\bottomrule
\end{tabular}
\label{tab:related_comparison}
\end{table}

\begin{figure*}[t]
    \centering
    \includegraphics[width=\linewidth]{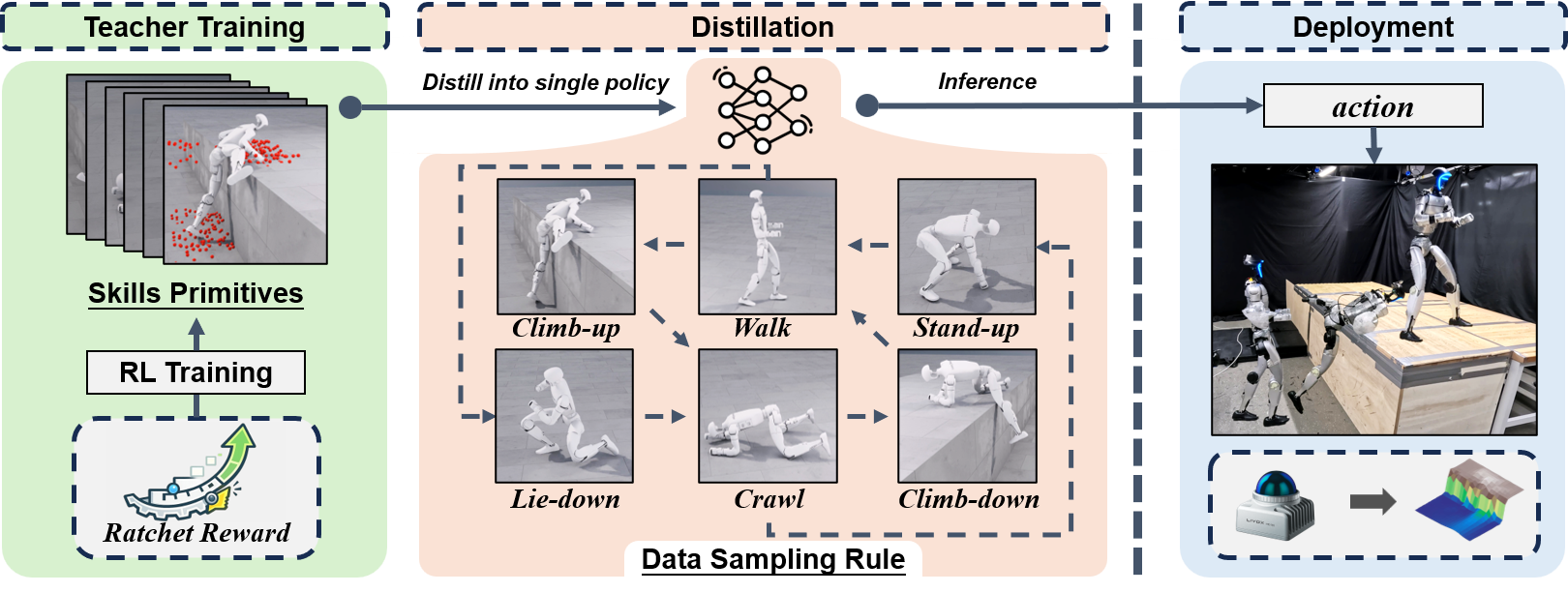}
    \caption{\textbf{Learning pipeline for high-platform traversal:} Teacher Training uses RL with the Ratchet Progress Reward, where a "best-so-far" task-space reference ensures genuine advancement by only rewarding states that strictly surpass historical progress. These skills are unified into a single context-aware policy through Distillation, using a ``divide-and-conquer'' Data Sampling Rule across distributed environments to cover the full distribution of maneuvers and transitions. In Deployment, the humanoid robot performs end-to-end traversal using LiDAR-based elevation mapping for terrain adaptation.}
    \label{fig:pipeline}
\end{figure*}

\subsection{Learning Feet-Based Locomotion}
Deep reinforcement learning (DRL) has substantially improved the robustness and agility of legged locomotion, with early successes on quadrupeds~\cite{tan2018sim, hwangbo2019learning, lee2020learning, rudin2022learning} and recent progress on humanoids.
Humanoids with learned controllers can walk and run in uneven terrain~\cite{radosavovic2024real, gu2024advancing, xue2025unified}, traverse sparse footholds~\cite{, wang2025beamdojo, ren2025vb, he2025attention}, and step or jump into elevated structures~\cite{zhuang2024humanoid, zhang2024whole, ben2025gallant}.
However, these methods primarily rely on \emph{feet-only} contacts, which requires large impulsive torques to reach high platforms, resulting in limited height (typically below $63\%$ of leg length).
In contrast, our system exploits whole-body, multi-contact coordination to distribute load and traverse platforms exceeding $114\%$ of leg length.

\subsection{Learning Humanoid Full-Body Maneuvers}
Recent work has started to learn individual full-body maneuvers such as stand-up~\cite{chen2025hifar, he2025learning, huang2025learning}. However, these methods often use task rewards that conflict with safety regularization and therefore require multi-stage pipelines or rely on heavy task-specific engineering (e.g., virtual-force shaping and carefully tuned curriculum for regularization and action rescaling), which has largely limited progress to relatively simple behaviors.
In contrast, our generalized \emph{ratchet progress} reward supports single-stage RL training for multiple contact-rich, goal-reaching maneuvers, while jointly optimizing task completion and safety regularization.

A complementary line of work learns human-like behaviors (e.g., dancing, walking, crawling, and jumping) by training policies to track human motions~\cite{cheng2024expressive, ji2024exbody2, fu2024humanplus, he2025hover, ze2025twist, chen2025gmt, liao2025beyondmimic} using dense imitation rewards~\cite{peng2018deepmimic}.
Building on these foundations, OmniRetarget~\cite{yang2025omniretarget} enables full-body climbing by preserving robot-scene contact relationships during retargeting.
However, motion-tracking approaches fundamentally rely on prerecorded trajectories and therefore require close alignment between the reference motion, the environment geometry, and the robot's initial state. This strongly limits adaptation to unseen terrain and initial conditions common in real deployment.
Motion generation models~\cite{tevet2022human, jiang2024harmon, li2025genmo} could in principle provide adaptive references, but generated motions are often physics-infeasible and do not explicitly reason about multi-contact feasibility and deployment constraints (e.g. torque limit, contact force, generation speed, perception gap) ; consequently, they are typically validated only in simulation~\cite{xu2025parc}.
In contrast, our policy is perceptive, reference-free, and deployable: it learns terrain-conditioned strategies that generalize across platform heights and initial poses, enabling autonomous traversal in diverse real-world environments.

\subsection{Policy Distillation for Legged Robots}
Teacher-student distillation is widely used to train deployable policies for legged robots~\cite{kumar2021rma, miki2022learning, fu2023deep, cheng2024extreme, yang2025agile, lin2025locotouch}. 
A common paradigm trains a teacher with privileged simulation information and distills it into a student that relies only on onboard observations (e.g., depth images or tactile signals) for deployment.
More recently, multi-expert distillation has been used to integrate terrain-conditioned skills into a single quadrupedal policy~\cite{zhuang2023robot, rudin2025parkour}, typically using DAgger-style~\cite{ross2011reduction} data aggregation, which we also adopt.
However, existing multi-skill distillation has largely focused on quadrupeds, where skills share similar feet-contact modes and transitions occur near a nominal walking posture.
In contrast, our teacher set spans heterogeneous humanoid behaviors, including full-body maneuvers and locomotion skills with substantially different state and action distributions (e.g., climbing, walking, crawling, and posture transitions).
This substantially increases the difficulty of both RL and distillation training: the teacher skills must be trained with compatible terminal-state distributions to enable safe and smooth concatenation, and appropriate teacher actions should be provided conditioned not only on terrain geometry and user commands, but also on the robot state and transition progress.


\section{High-Platform Traversal Policy Learning}

Our goal is to learn a perceptive humanoid policy that can robustly traverse extremely high platforms in the real world.
As introduced in Sec.~\ref{sec:intro}, such traversal requires multiple terrain-conditioned behaviors: four full-body maneuvers (\emph{climb-up, climb-down, stand-up, lie-down}) and two standard locomotion skills (walking, crawling).
To handle this diversity, we adopt a two-stage learning pipeline~\cite{, yang2025agile, lin2025locotouch} as shown in Fig. \ref{fig:pipeline}.
First, we develop a unified RL training framework to learn the four full-body maneuver policies with LiDAR perception (Sec.~\ref{sec:rl_training}).
To enable efficient learning of these contact-rich goal-reaching tasks, we introduce a generalized \emph{ratchet progress} reward that provides dense supervision while supporting exploration under strong safety regularization (Sec.~\ref{sec:task_reward}).
Second, we distill all six policies into a unified single policy that autonomously selects and transitions between behaviors based on perception, enabling end-to-end high-platform traversal (Sec.~\ref{sec: distillation}).

\subsection{Ratchet Progress Reward for Humanoid Maneuvers}
\label{sec:task_reward}

\subsubsection{Task Definition} \mbox{}

We model the full-body, contact-rich humanoid maneuver as a goal-reaching task, where success is defined by satisfying a terminal condition rather than tracking a reference trajectory.
Let $s_t$ denote the robot state at timestep $t$. 
We define a \textit{task state} $x_t = \phi(s_t)$, where $\phi(\cdot)$ extracts a minimal set of variables needed to evaluate task completion.
For each maneuver, we specify a target task state $x^g$ and declare success when $x_t \ge x^g$, where $\ge$ denotes the ordering induced by the task metric.

To instantiate the four maneuver objectives (Tab.~\ref{tab:task_definition}), we use the following notation.
$p_{CoM}$, $p_{head}$, and $p_{LB}$ denote the positions of the full-body center of mass, head, and lower body, respectively.
We use $h$ and $x$ to denote environment- and pose-dependent thresholds, such as the platform edge height $h_{edge}$ or the nominal standing head height $h_{head}^{stand}$.
For standing stability, we define a balance margin $d_{bal} = \| p_{CoM} - \bar{p}_{feet} \|$ where $\bar{p}_{feet}$ is the geometric center of the feet.
These definitions yield concise, task-specific terminal conditions while keeping $x_t$ low-dimensional and easy to compute online.

\begin{table}[h]
\centering
\setlength{\tabcolsep}{3pt}
\caption{Task Definition of Four Goal-Reaching Maneuvers}
\footnotesize
\begin{tabular}{lcc}
\toprule
\textbf{Task} & \textbf{Task State ($x_t$)} & \textbf{Target Task State ($x^{g}$)} \\
\midrule
\textbf{Climb-up} 
& $p_{CoM}, p_{LB}$ 
& $p_{LB}^{(z)} > h_{edge} \land p_{CoM}^{(x)} > x_{edge}$ \\ [4pt]
\textbf{Climb-down} 
& $p_{CoM}, p_{LB}$ 
& $p_{LB}^{(z)} < h_{LB}^{stand} \land p_{CoM}^{(x)} < x_{edge}$ \\ [4pt]
\textbf{Stand-up} 
& $p_{head}^{(z)}, d_{bal}$ 
& $p_{head}^{(z)} > h_{head}^{stand} \land d_{bal} < \delta$ \\ [4pt]
\textbf{Lie-down} 
& $p_{CoM}, p_{head}$ 
& $p_{CoM}^{(z)} < h_{CoM}^{prone} \land p_{head}^{(z)} < h_{head}^{prone}$ \\ [2pt]
\bottomrule
\end{tabular}
\label{tab:task_definition}
\end{table}

\subsubsection{Ratchet Progress Reward} \mbox{}

These goal-reaching maneuvers do not admit a phase-invariant predefined reference, such as a consistent velocity or contact pattern for command-conditioned locomotion.
To provide a meaningful reference at every timestep without prescribing a motion template, we introduce a \emph{self-updating task-space reference} that records the best progress achieved so far along the trajectory.
This best-so-far task state at timestep $t$ is defined as:
\begin{equation}
x_t^* = \max(x_0, x_1, \ldots, x_{t-1})
\label{eq:best_so_far_reference}
\end{equation}
which can be updated online via $x_t^* = \max(x_{t-1}^*, x_{t-1})$, with $x_0^*=x_0$.
Intuitively, $x^*_t$ tracks the current frontier of task-space progress demonstrated by the agent.

Based on the best-so-far task-space reference, we define a binary \textit{ratchet progress} task reward:
\begin{equation}
r_t =
\begin{cases}
0, & \text{if } x_t > x_t^*,
\\-1, & \text{otherwise.}
\end{cases}
\label{eq:ratchet_reward}
\end{equation}
That is, the agent receives no penalty only when it \emph{strictly surpasses} its historical best, and is penalized otherwise.
Although simple, the above construction is tailored to contact-rich, goal-reaching maneuvers, with three key properties.

\noindent\textbullet\hspace{0.6em}\textbf{Dense task-aligned supervision.}
The reward is evaluated at every timestep to penalize any failure to exceed the best-so-far progress.
This provides a dense signal that keeps exploration within task-relevant behaviors, which is essential for contact-rich maneuvers where terminal-only rewards are too sparse to discover feasible contact sequences.

\noindent\textbullet\hspace{0.6em}\textbf{Velocity-free progress enables exploration and deployment.}
Because the reward depends only on \emph{whether} progress improves but not \emph{how much}, it does not encourage rushing in task space.
This supports (i) \emph{patient, contact-aware exploration}, allowing the robot to hold intermediate supports until necessary contacts become stable (e.g., during climb-up, keep one leg grounded until the other stably lands on the platform; during stand-up, hold torso ascent until limbs become load-bearing), and (ii) \emph{effective regularization} of impact/torque/force can enforce safe motions without having to counteract a velocity-driven task incentive.

\noindent\textbullet\hspace{0.6em}\textbf{History dependence prevents retracing exploits.}
Incremental criteria such as $(x_t > x_{t-1})$ can be gamed by oscillating backward and forward.
In contrast, our historical best criteria $(x_t > x_t^*)$ ensures optimization with genuine advancement toward the goal.

\subsection{Learning Perceptive Full-Body Maneuvers}
\label{sec:rl_training}
In this section, we present our RL framework for learning the four full-body maneuver skills.
For brevity, we omit the training details of the two standard locomotion skills, as they follow a conventional velocity-tracking formulation.

\subsubsection{RL Training Environment} \mbox{}

\textbf{State, Observation, and Action.}
We train each maneuver as a single-skill policy in a Markov Decision Process (MDP).
The observation space includes robot proprioception $s_t^{\text{proprio}} \in \mathbb{R}^{64}$ (gravity vector, base angular velocity, and joint positions/velocities), the previous action $a_{t-1}\in\mathbb{R}^{29}$, the task state $x_t$, and optionally a local elevation map $m_t \in \mathbb{R}^{441}$ at $0.05\,\text{m}$ resolution covering a $1\times1\,\text{m}^2$ area.
All policies take a 5-step history of $(s_t^{\text{proprio}}, a_{t-1})$ to capture short-term dynamics.
Because the climbing skills must perceive the platform geometry, the \emph{climb-up} and \emph{climb-down} policies additionally take $m_t$ as input.
All policies output target joint positions $a_t \in \mathbb{R}^{29}$, which are tracked by a low-level PD controller.

Our task reward depends on the best-so-far task state $x_t^*$, which evolves over time and is not included in the instantaneous robot state. This creates history dependence that can impair value estimation if the critic observes only $s_t$. To address this, we provide $x_t^*$ as additional input to the critic.

\textbf{Simulation Environment.}
To learn adaptive behaviors, we extensively randomize the terrain configuration and initial conditions.
The platform height is sampled from $[0.55\,\mathrm{m},\,0.85\,\mathrm{m}]$.
For \emph{climb-up}, the initial distance from the robot base to the vertical surface and the initial yaw angle are sampled from $[0.15\,\mathrm{m},\,0.35\,\mathrm{m}]$ and $[-60^\circ,\,60^\circ]$; for \emph{climb-down}, they are sampled from $[0.30\,\mathrm{m},\,0.45\,\mathrm{m}]$ and $[-75^\circ,\,75^\circ]$.
To improve sim-to-real transfer, we apply a comprehensive suite of domain randomization following~\cite{liao2025beyondmimic}.
We further apply symmetric augmentation~\cite{mittal2024symmetryconsiderationslearningtask} to reduce handedness bias and improve generalization across approach angles.

\textbf{Initial Posture Sampling.}
Since single-skill policies are executed sequentially during distillation, their initial-state distributions must encompass the terminal states generated by preceding skills. As illustrated in Fig.~\ref{fig:pipeline}, transitions predominantly occur around two canonical postures: standing (start of walking, climb-up, and lie-down; end of walking, climb-down, and stand-up) and prone (start of crawling, climb-down, and stand-up; end of crawling, climb-up, and lie-down). We define nominal joint configurations for these postures as $q_{\text{stand}}$ and $q_{\text{prone}}$, respectively.

For each skill, initial joint angles are sampled by perturbing the corresponding nominal starting posture, ensuring that training begins from physically plausible states that are compatible with upstream transitions. To enable seamless switch-out between skills, we additionally shape the terminal behavior of full-body maneuvers toward the nominal ending posture using a terminal-pose reward (Sec.~\ref{sec:reward_design}). If the reachable terminal-state distribution of a skill is not fully contained within the initial-state distribution of its successor, we subsequently retrain the successor skill while augmenting its initial-state distribution to cover all possible terminal configurations produced by its predecessors.

\subsubsection{Reward Design} \mbox{}
\label{sec:reward_design}

We define the total reward as the sum of five components:
\begin{equation}
    r = r_{\text{alive}} + r_{\text{reg}} + r_{\text{force}} + r_{\text{task}} + r_{\text{tp}}
\end{equation}

The first three terms $r_{\text{alive}}$, $r_{\text{reg}}$, and $r_{\text{force}}$, are shared across all maneuvers.
$r_{\text{alive}}$ is a constant survival bonus that discourages early termination.
$r_{\text{reg}}$ aggregates standard regularization penalties that promote smooth and energy-efficient motions.

Because full-body maneuvers involve frequent terrain contacts beyond the feet, limiting impact is critical for safe deployment.
We therefore include a contact-force penalty $r_{\text{force}}$ that grows rapidly once contact forces exceed a safe threshold:
\begin{equation}
    r_{\text{force}} = -\left( \exp \left( \alpha \cdot \max(0, F_t - F_{limit}) \right) - 1 \right)
\end{equation}
where $F_t$ is the maximum contact force at timestep $t$, $F_{limit}$ is a safety threshold, and $\alpha>0$ controls the penalty scale. Specially, we set $F_{limit}=0$ for the head link, since even light head contact can destabilize the head-mounted LiDAR and severely degrade perception.

The remaining two terms, $r_{\text{task}}$ and $r_{\text{tp}}$, are task-specific but require only minimal specification.
$r_{\text{task}}$ is the ratchet progress reward introduced in Sec.~\ref{sec:task_reward}, which drives goal completion.
$r_{\text{tp}}$ encourages a desired terminal posture to facilitate reliable behavior transitions.
It is activated \emph{only} after the goal is reached and within the final second of an episode:
\begin{equation}
    r_{\text{tp}} = \mathbb{I}_{(t > H - 1s)} \cdot \mathbb{I}_{(\text{goal reached})} \cdot \exp(-\beta \| q_t - q_{tar} \|^2)
\end{equation}
where $\mathbb{I}$ is the indicator function, $H$ is the episode duration, $q_t$ and $q_{tar}$ denote the current and desired terminal joint angles of the robot, and $\beta>0$ is a scale parameter.

\subsubsection{Robust Perception via Elevation Mapping}
\label{sec:perception}
\begin{figure}[t]
    \centering
    \includegraphics[width=\linewidth]{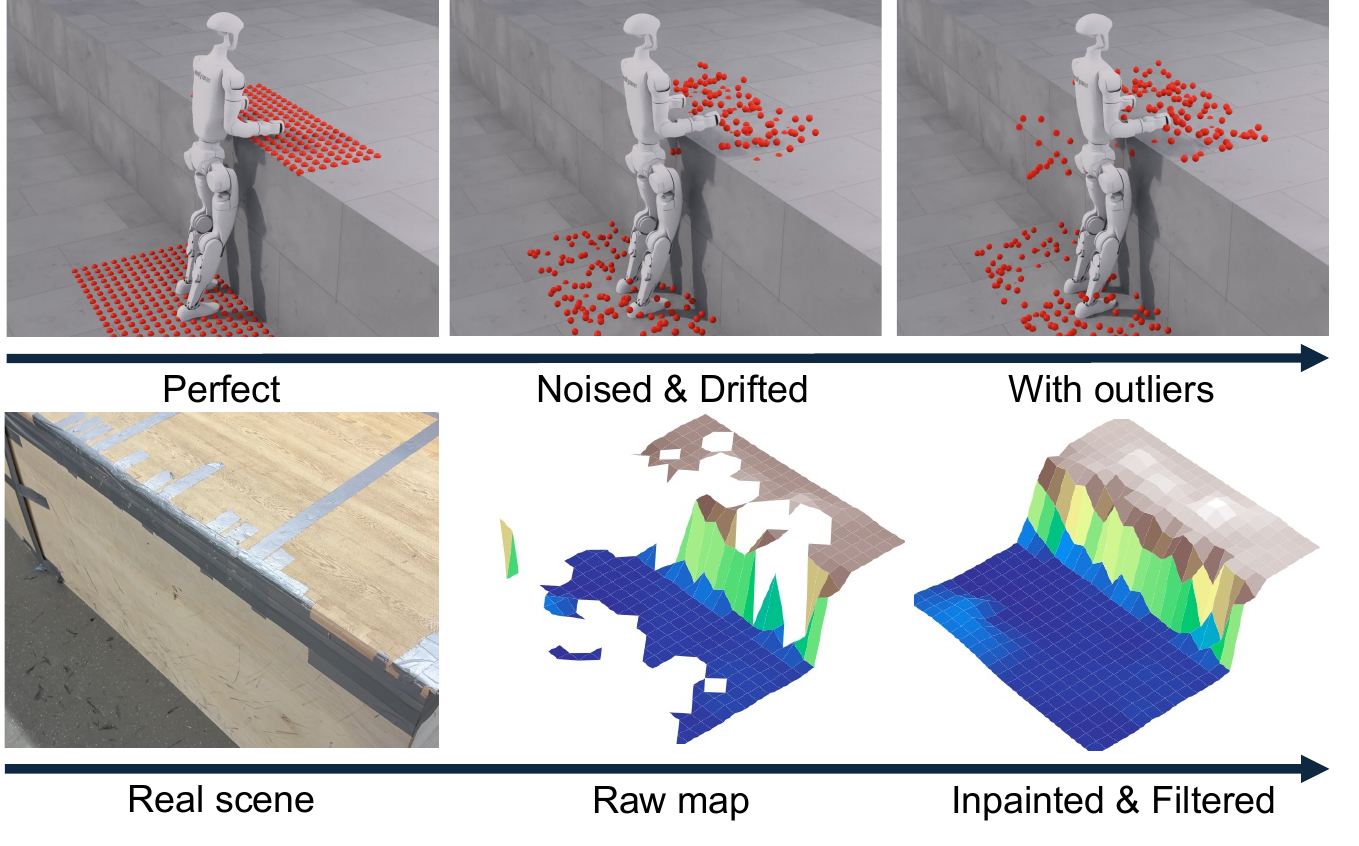}
    \caption{The sim-to-real gap in LiDAR mapping is addressed through a dual approach that combines artifact modeling in simulation with real-world post-processing.}
    \label{fig:lidar}
\end{figure}

We implement a LiDAR-based elevation mapping pipeline following prior humanoid locomotion work~\cite{long2025learning}. 
However, dynamic maneuvers introduce perceptual degradation: rapid accelerations and contact-induced disturbances accumulate localization drift; self-occlusion from the robot’s limbs produces spurious point clusters; extreme body configurations limit field of view; and probabilistic elevation fusion yields uncertain or missing measurements. To mitigate the sim-to-real perception gap induced by these, we adopt a dual strategy (Fig.~\ref{fig:lidar}):

\begin{itemize}
    \item \textbf{Simulation Artifact Modeling:} During training, we explicitly inject three classes of perceptual artifacts to improve robustness: per-cell Gaussian noise to emulate mapping uncertainty and self-scanning effects, spatial offsets to simulate localization drift, and synthetic outlier clusters to reproduce spurious obstacle artifacts.
    \item \textbf{Real-World Post-Processing:} We apply a spatial outlier filter to suppress high-variance noise clusters and an inpainting algorithm~\cite{bradski2008learning} to reconstruct missing elevation regions, providing the policy with a structurally coherent terrain representation.
\end{itemize}

\begin{figure*}[t]
    \centering
    \includegraphics[width=1\linewidth]{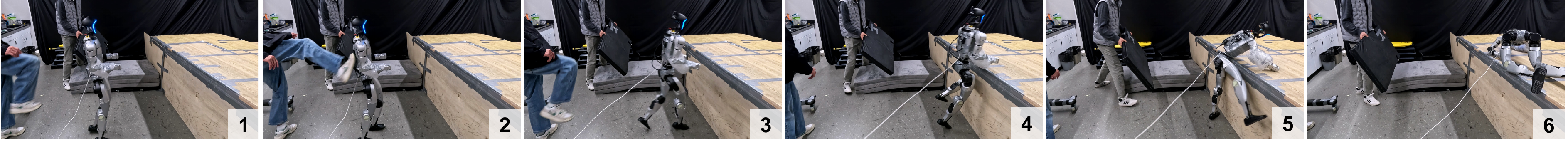}
    \caption{The robot regains balance and climbs up the high platform after being heavily kicked. }
    \label{fig:kick}
    \vspace{-10pt}
\end{figure*}

\subsection{Policy Distillation for Skill Integration}
\label{sec: distillation}

To obtain a unified, context-aware controller from our diverse skill library, we adopt a distillation pipeline inspired by~\cite{lin2025locotouch}. Specifically, we first leverage Behavior Cloning (BC)~\cite{torabi2018behavioral} to pretrain a student policy, and then refine it using DAgger~\cite{ross2011reduction} to improve robustness and distributional coverage.
We dynamically select the appropriate teacher based on the robot's state and and commanded velocity, optimizing the policy using the Mean Squared Error (MSE) between the student and teacher actions.

\subsubsection{Construction of Data Distribution}
A standard sequential sampling strategy~\cite{zhuang2023robot} is unsuitable for high-platform traversal distillation, which involves multiple substantially different behaviors. Rollouts starting from the traversal initial state produce highly imbalanced data. For example, predominantly walking data in early stages, or no climb-down data until a successful climb-up occurs.
To obtain a more balanced dataset across all teacher policies while covering possible skill transitions, we introduce a ``divide-and-conquer'' strategy: each training environment is assigned either to a single core skill or to a combination of two consecutive skills. More detail for the environment definition is provided in the Appendix.


\subsubsection{Data Augmentation Strategy}
We incorporate the full suite of configurations utilized during teacher training, including domain randomization, physical perturbations, and perception artifacts. The first skill in each sub-environment is initialized from a broader range of states to maximize state-space coverage for the student policy. Additionally, we apply action noise and symmetry augmentation during the distillation process to improve the robustness of the unified policy. 



\section{Experiments}

In this section, we present a series of qualitative and quantitative evaluations to address the following questions:
\begin{enumerate} 
    \item Does the proposed system enable context-aware traversal via coordinated skill execution
 (Sec.~\ref{sec:system_perf})?
    \item Do the learned policies demonstrate robustness and adaptability to environmental variations (Sec.~\ref{sec:single_skill_perf})?
    \item How does the proposed ratchet progress reward facilitate the acquisition of full-body maneuvers (Sec.~\ref{sec:reward_perf})?
\end{enumerate}

Experiments are conducted on a 29-DoF Unitree G1 humanoid robot in both simulation and real-world settings.


\subsection{Performance of Context-Aware Continuous Traversal}
\label{sec:system_perf}

\textbf{The system enables continuous traversal via coordinated skills.}
To evaluate context-aware traversal across high platforms, we design three challenging simulation courses requiring long-horizon execution without resets (Fig.~\ref{fig:teaser}). These courses—\emph{Continuous Traversing}, \emph{Ascending}, and \emph{Descending}—combine acyclic maneuvers (e.g., climbing, standing up, lying down) with periodic gaits (e.g., walking) into cohesive traversal sequences. Successful completion requires the policy to autonomously determine skill transitions based on terrain context. The robot is commanded via velocity inputs, while standing up and lying down are triggered by the user.

To assess system-level robustness, we introduce environmental perturbations and perception artifacts during evaluation, including LiDAR degradation and state-estimation drift. Despite these disturbances, the robot maintains stable traversal across varying terrain geometries. Over 1,000 trials with predefined command sequences, the policy achieves a 95.4\% success rate.

\textbf{Zero-shot transfer to long-sequence real-world deployment.}
As shown in Fig.~\ref{fig:teaser}, we validate the learned policy on hardware via zero-shot sim-to-real transfer. The robot autonomously coordinates walking, climb-up, stand-up, lie-down, and climb-down skills to traverse a 0.8\,m platform in a continuous full-loop sequence. Additional results involving consecutive full-loop traversals are provided in the Appendix. During deployment, the robot exhibits context-aware motor strategies: it switches to climb-up when approaching the platform and selects different lead legs depending on the approach angle, while the descending phase follows a similar adaptive pattern. These behaviors demonstrate the policy’s ability to modulate skill execution based on the surrounding physical context.

\textbf{Robust skill transitions under severe perturbations.}
Beyond nominal traversal, the system remains effective under strong external disturbances. As shown in Fig.~\ref{fig:kick}, the robot is heavily kicked from behind while approaching the platform, causing a stumble and unintended contact. Despite being pushed into a near-failure state, it rapidly adapts by adjusting its gait and switching the pivoting leg to stabilize and initiate climbing. This behavior demonstrates that the distilled policy executes skill transitions even from near-failure states, leveraging environmental contact to regain balance. These results indicate that distillation transfers teacher robustness into a unified policy capable of context-dependent modulation.


\subsection{Robustness and Adaptability to Environmental Variations}
\label{sec:single_skill_perf}

\begin{figure}[t]
    \centering
    \includegraphics[width=\linewidth]{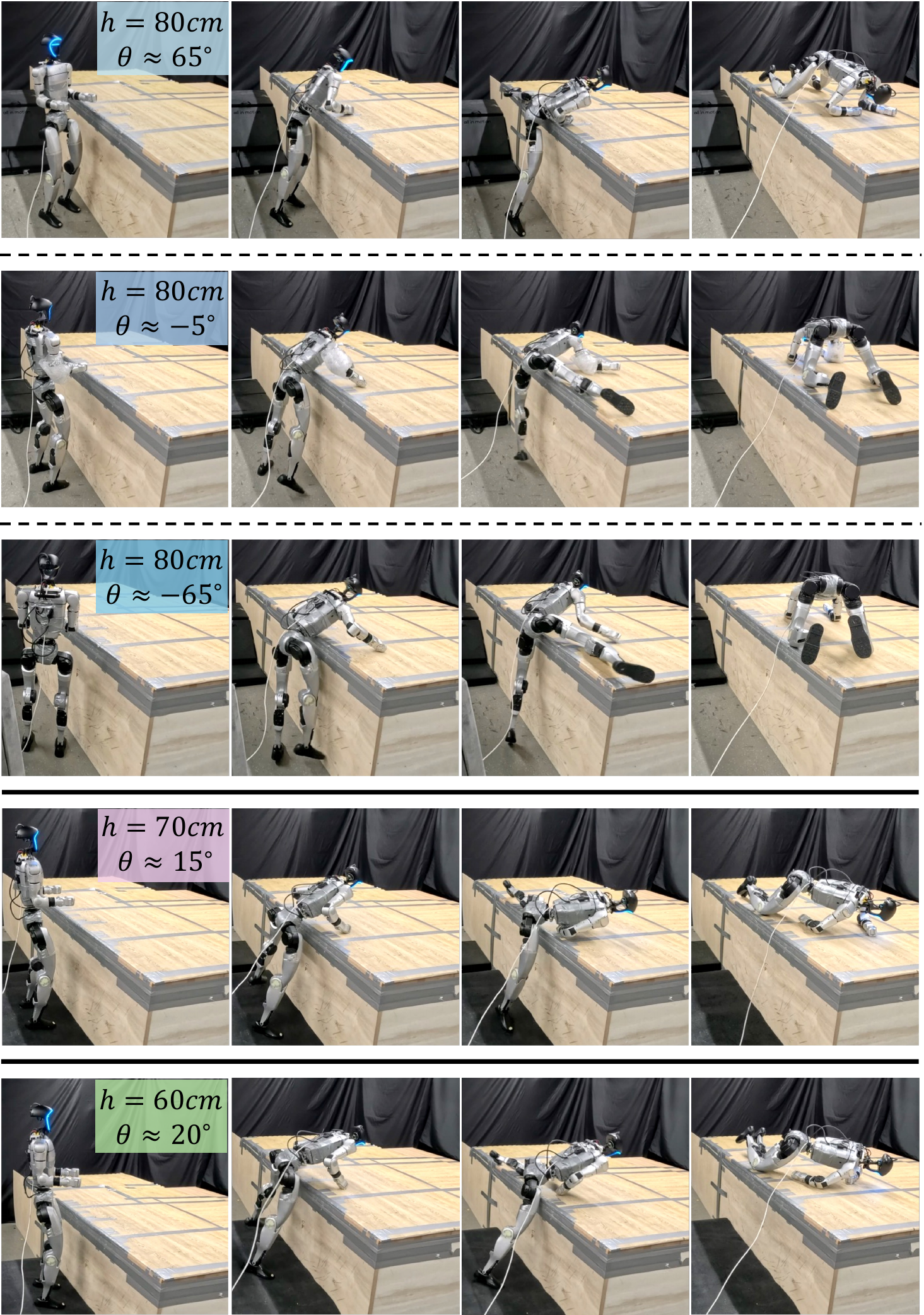}
    \caption{Real-world adaptation of the climb-up policy to varying platform heights (0.6–0.8,m) and approach angles ($\theta \in [-65^\circ, 65^\circ]$). The policy exhibits coordinated whole-body behaviors and reliable zero-shot sim-to-real transfer, even in extreme out-of-distribution cases.}
    \vspace{-10pt}
    \label{fig:climbup}
\end{figure}

\begin{table}[t]
\centering
\caption{Comparative Success Rates of Single Skill in Sim and Real. SR: Success Rate; S/T: Success / Trials; M.C.F.: Max Contact Force; $H$ (m): platform height; $A$ ($^\circ$): approach angle relative to the platform-edge normal.}
\label{tab:success_rates}
\footnotesize 
\setlength{\tabcolsep}{3pt}
\begin{tabular}{@{}l cc cccc@{}}
\toprule
& \multicolumn{2}{c}{\textbf{Simulation}} & \multicolumn{4}{c}{\textbf{Real World}} \\
\cmidrule(lr){2-3} \cmidrule(lr){4-7}
\textbf{Task} & \textbf{S.R. (\%)} & \textbf{M.C.F. (N)} & \textbf{$H$ (m)} & \textbf{$A$ ($^\circ$)} & \textbf{S/T} & \textbf{S.R. (\%)} \\ 
\midrule
\multirow{9}{*}[-1ex]{Climb-up} & \multirow{9}{*}{98.8} & \multirow{9}{*}{\begin{tabular}{c}
    $638$ \\
    $\pm\,479$
  \end{tabular}} & 0.6 & $[-45, -15]$ & 5/5 & \multirow{9}{*}{97.8} \\
 & & & 0.6 & $[-15, +15]$ & 5/5 & \\
 & & & 0.6 & $[+15, +45]$ & 5/5 & \\ \cmidrule(lr){4-6}
 & & & 0.7 & $[-45, -15]$ & 4/5 & \\
 & & & 0.7 & $[-15, +15]$ & 5/5 & \\
 & & & 0.7 & $[+15, +45]$ & 5/5 & \\ \cmidrule(lr){4-6}
 & & & 0.8 & $[-45, -15]$ & 5/5 & \\
 & & & 0.8 & $[-15, +15]$ & 5/5 & \\
 & & & 0.8 & $[+15, +45]$ & 5/5 & \\
\midrule

Climb-down & 99.9 & {\begin{tabular}{c}
    $754$ \\
    $\pm\,241$
  \end{tabular}} & 0.8 & $[-45, +45]$ & 5/5 & 100.0 \\
Stand-up   & 99.5 & \begin{tabular}{c}
    $632$ \\
    $\pm\,222$
  \end{tabular}  & --  & --           & 5/5 & 100.0 \\
Lie-down   & 100.0  & {\begin{tabular}{c}
    $576$ \\
    $\pm\,125$
  \end{tabular}}  & --  & --           & 5/5 & 100.0 \\
\bottomrule
\end{tabular}
\end{table}

\textbf{Robustness evaluation of individual full-body maneuvers.}
To quantify the robustness of each full-body maneuver, we first evaluate the specialized teacher policies across a diverse set of challenging scenarios in simulation. Each task is assessed over 1,000 independent trials, while maintaining identical terrain distributions, domain randomizations, and pose initializations as those used during training. We adopt Success Rate (SR) as the primary performance metric, and additionally report the maximum contact force (M.C.F.) to assess safety. As summarized in Tab.~\ref{tab:success_rates}, all teacher policies achieve near-perfect success rates in simulation, while maintaining contact forces within safe limits. These results validate the effectiveness of the proposed contact-force regularization across varied terrain geometries and initial conditions.

\textbf{Generalization across real-world configurations.}
We further evaluate the robustness and adaptability of the learned policies on hardware across multiple real-world configurations. For the climb-up task, we vary the platform height from 0.6\,m to 0.8\,m and the approach angle from $-45^\circ$ to $45^\circ$, as summarized in Tab.~\ref{tab:success_rates}. Across these in-distribution configurations, the policy consistently achieves high success rates, including at a platform height of 0.8\,m, which corresponds to approximately 114\% of the robot’s leg length.

In addition to these settings, we evaluate the policy under more extreme approach angles of up to $\pm 65^\circ$, which lie outside the training distribution. As illustrated in Fig.~\ref{fig:climbup}, both in-distribution and out-of-distribution cases are shown. In these cases, the robot demonstrates strong adaptability by modifying its whole-body strategy according to the approach geometry, reorienting its torso toward the platform and leveraging full-body motion to initiate the climb rather than executing a fixed or naive forward reach.

Beyond climb-up, we also evaluate other full-body maneuvers, including climb-down, stand-up, and lie-down, on hardware. As reported in Tab.~\ref{tab:success_rates}, all evaluated skills achieve a 5/5 success rate, indicating that the robustness and adaptability of the learned behaviors extend beyond a single maneuver.

\textbf{Robustness to Varying Contact Properties.}
We evaluate the climbing policy under out-of-distribution contact conditions by placing a soft vinyl–foam mat on the target platform (see Appendix). This introduces unseen compliance and friction properties compared to the rigid training surfaces. The robot successfully climbs onto the soft surface while maintaining stability consistent with behavior observed on rigid platforms.

\textbf{Symmetry augmentation facilitates balanced behaviors.}
With symmetry augmentation, the policy converges to a balanced strategy rather than a biased handedness. During climb-up, the lead leg is selected dynamically based on the robot’s relative heading to the platform. Such balance is critical in practice, as asymmetric biases restrict the feasible workspace and degrade climbing performance.


\begin{figure}[t]
    \centering
    \includegraphics[width=1\linewidth]{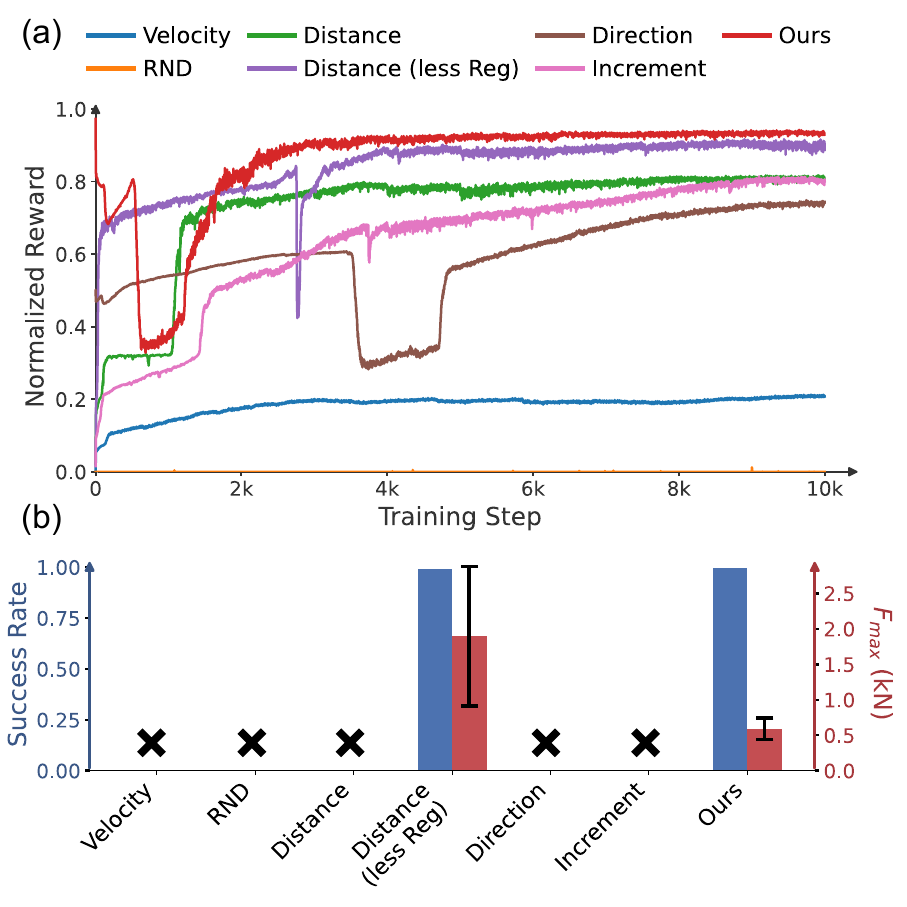}
    \caption{Comparative analysis of reward formulations. (a) Learning curves for normalized task rewards. (b) Success rate (blue) and maximum contact force (red) of the trained policies. }
    \vspace{-6pt}
    \label{fig:training_curve}
\end{figure}
\vspace{-5pt}

\subsection{Skill Acquisition via Progress Rewards}
\label{sec:reward_perf}
To evaluate the efficacy of progress-based rewards in learning adaptive full-body maneuvers, we benchmark the training of the Climb-up skill against several baseline formulations. In each baseline, the progress-based reward is replaced by alternative task rewards while maintaining identical weights and hyperparameters. 
\begin{enumerate}
    \item Velocity: Employs a standard velocity-tracking objective aimed at matching a  torso velocity command\cite{zhuang2024humanoidparkourlearning,long2025learning,he2025attention}. This command is defined in the world frame and oriented forward toward the platform.

    \item RND (Random Network Distillation~\cite{burda2018explorationrandomnetworkdistillation}): Utilizes intrinsic rewards generated via RND to incentivize exploration, combined with a sparse task reward upon task completion.~\cite{pmlr-v229-schwarke23a, zhang2024wococolearningwholebodyhumanoid}. 

    \item Distance: Penalizes the distance to the target, encouraging the agent to minimize this gap at every timestep.

    \item Distance (less Reg.): Follows the same distance-minimization objective as above but with significantly lower regularization penalties.

    \item Direction: Rewards any base velocity in the direction towards the goal,  while penalizing small velocity to prevent stalling~\cite{hoeller2024anymal,cheng2023extremeparkourleggedrobots}.
    \item Increment: Rewards the difference between the previous and
    the current distance of the system state from the goal state~\cite{openai2019solvingrubikscuberobot}.
\end{enumerate}

\begin{figure}[t]
    \centering
    \includegraphics[width=\linewidth]{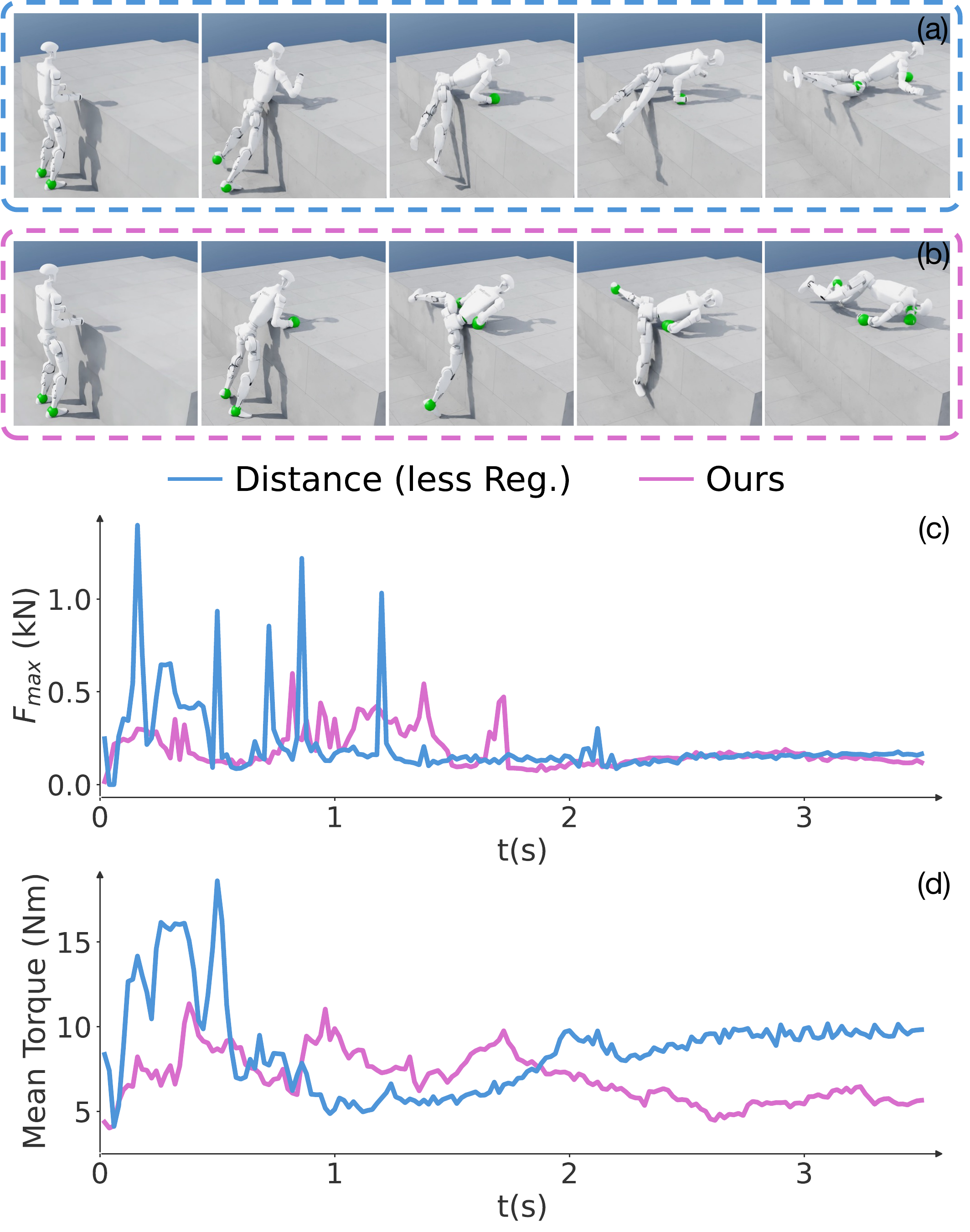}
    \caption{Comparison against the baseline. (a)(b) Keyframes of policy trained separately with distance-based reward / our proposed reward. The actual contact points at each timestep are visualized with the green sphere. (c) Max contact force over body parts w.r.t. time. (d) Mean joint torque w.r.t time. }
    \label{fig:comp_baseline_traj}
    \vspace{-0.3cm}
\end{figure}

The details of the baseline training configurations are shown in Appendix. We report the learning curves for task rewards, alongside the success rates and maximum contact forces for each formulation. To ensure statistical significance and assess robustness, each policy is evaluated over 1,000 independent trials across randomized environment configurations. These results are summarized in Fig~\ref{fig:training_curve}.

Baseline 1 fails to complete climb-up and remains stuck at the platform edge. The velocity-tracking objective over-constrains motion to a fixed forward speed, preventing the discovery of adaptive velocity modulation required to negotiate the edge. Although the reward is partially optimized, task completion remains unsuccessful.

Baseline 2 fails due to the absence of structured guidance for precise multi-stage coordination. Although curiosity-driven exploration encourages diverse behaviors, it lacks a directional gradient toward task completion, often resulting in exploration of physically irrelevant states. Without a dense shaping signal such as \rewardname, the probability of discovering the specific maneuver sequence required to trigger sparse success within the training budget is negligible.


The distance-minimization objective induces a strong velocity bias, encouraging rapid target approach to maximize cumulative returns. This conflicts with contact-force regularization, preventing Baseline 3 from achieving a balance between task success and physical safety. Reducing regularization in Baseline 4 enables goal reaching but produces aggressive “full-body jumping” behaviors characterized by impulsive contacts and excessive joint torques (Fig.~\ref{fig:comp_baseline_traj}). Such solutions are physically infeasible for real-world deployment. In contrast, the proposed ratchet progress-based reward yields sustained whole-body coordination with significantly reduced peak forces.

Baselines 5 and 6 converge to a local optimum at the platform edge, where the agent exhibits a repetitive "back-and-forth" motion—abruptly retreating only to slowly advance again. Because these rewards depend only on instantaneous velocity direction, forward progress can be accumulated without committing to the full maneuver. The agent therefore learns to cyclically reset its position to maximize reward accumulation rather than execute the climb.

Collectively, these results expose a fundamental limitation of instantaneous or goal-distance-based objectives: they either induce unsafe velocity bias or admit degenerate local optima. In contrast, the history-dependent ratchet progress reward produces sustained whole-body coordination, modulating contact locations and force distribution to achieve stable climb-up behaviors with substantially reduced peak forces suitable for hardware deployment.


\section{Conclusion}
\label{sec:conclusion}
We presented \nickname, a learning system for adaptive high-platform traversal on humanoid robots, targeting extreme ledges where jumping becomes unsafe and actuator-limited. Our approach trains six terrain-conditioned skills (four contact-rich maneuvers and two cyclic locomotion skills) and distills them into a single perceptive policy that autonomously selects behaviors and transitions from LiDAR-based elevation maps.
Central to our method is a generalized ratchet progress reward for goal-reaching maneuvers. By tracking best-so-far task progress and penalizing non-improving steps, it provides dense, velocity-free supervision that enables efficient learning under strong safety regularization and avoids retracing failure modes.
Experiments on a 29-DoF Unitree G1 humanoid demonstrate zero-shot sim-to-real traversal of $0.8\,\text{m}$ platforms ($\approx 114\%$ of leg length), with robust adaptation to platform height and initial pose and smooth multi-skill transitions.

\section*{Acknowledgment}

This work was partially supported by the National Science Foundation under Grant CAREER CNS-2047454. The authors would also like to thank Yuxiang Yang from Google DeepMind for valuable advice and discussions, and Boda Huo and Mingyang Yu from the CMU SafeAI Lab for their assistance with the experiments.

\newpage


{
\bibliographystyle{IEEEtran}
\bibliography{references}
}

\clearpage


\section{Appendix}


\subsection{Effectiveness of Ratchet Progress Reward}

\begin{figure}[h]
    \centering
    \includegraphics[width=\linewidth]{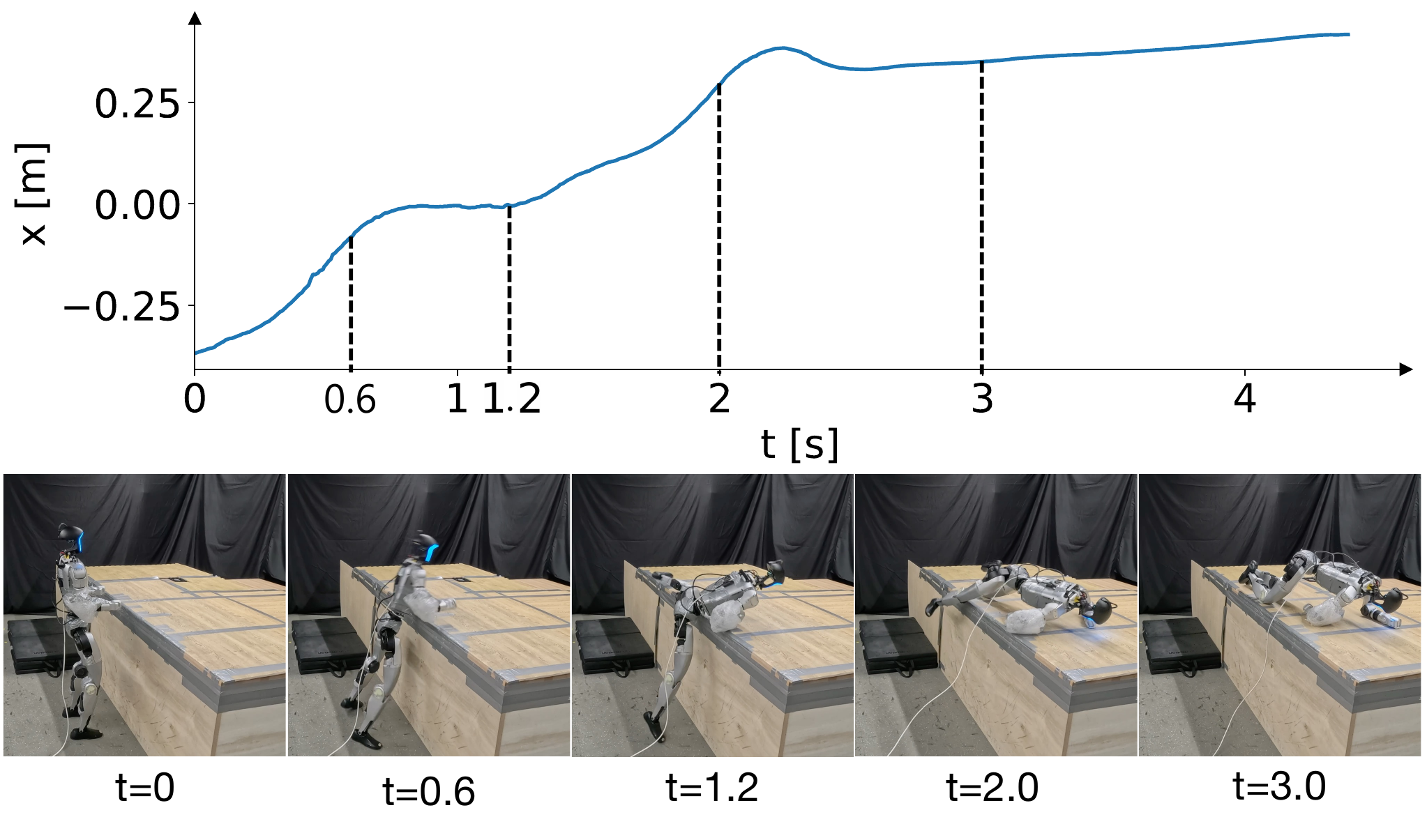}
    \caption{Trajectory of the robot's torso relative to the platform edge. The top plot shows the horizontal displacement $x(t)$ over time, while the bottom sequence illustrates the corresponding climbing up motion at key timestamps.}
    \label{fig:mocap_traj}
\end{figure}

We deploy our policy on hardware and record the torso horizontal displacement ($x$-direction) using a motion capture system (MoCap). We do not deploy the baseline policy due to the high risk associated with its excessive contact forces and impulsive movements. As illustrated in Fig.~\ref{fig:mocap_traj}, the measured trajectory reveals two characteristic properties of functional climb-up behavior: monotone task progress with contact-induced holds. After the initial approach and hand placement ($0$--$0.7,\text{s}$), the trajectory exhibits a pronounced plateau centered around $t\approx 1.0,\text{s}$. This stagnation phase is functionally necessary: the torso remains near the platform edge while the robot lifts and securely places the lead leg. Once this contact is established, the torso resumes forward progression ($t>1.2,\text{s}$), driven by coordinated forces from the hands and the newly established foothold. The emergence of this deliberate pause highlights the event-driven nature of contact-rich maneuvers and indicates that our reward formulation learns to prioritize kinematic feasibility and stability, rather than simply minimizing distance to the goal.

\subsection{Extended Experiments on Context-Aware Traversal}

\begin{figure}[h]
    \centering
    \captionsetup[subfigure]{skip=2pt} 

    \begin{subfigure}{\columnwidth}
        \centering
        \includegraphics[width=1.0\linewidth]{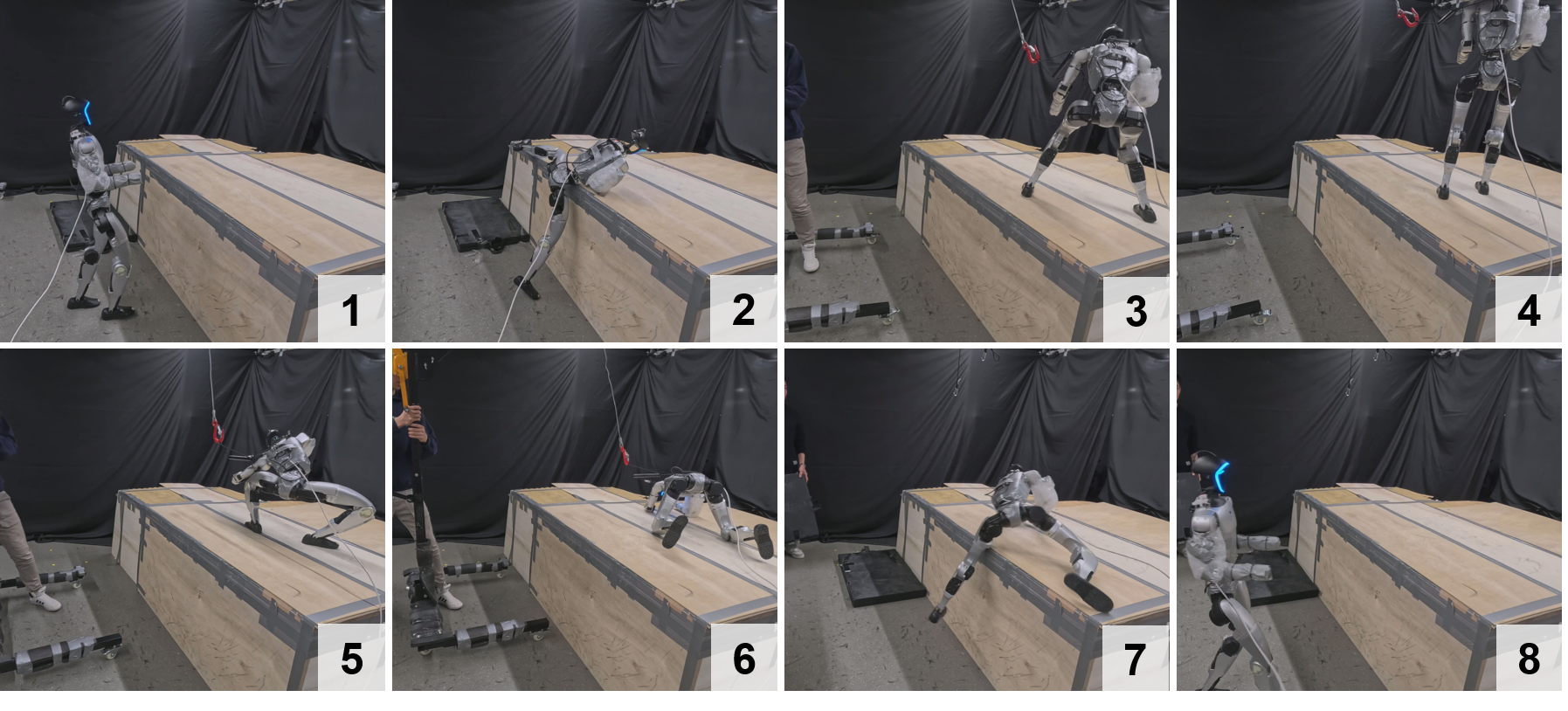}
        \caption{Robot executing two consecutive full-skill cycles.}
        \label{fig:cycle2}
    \end{subfigure}
    
    \vspace{0.1cm} 

    \begin{subfigure}{\columnwidth}
        \centering
        \includegraphics[width=1.0\linewidth]{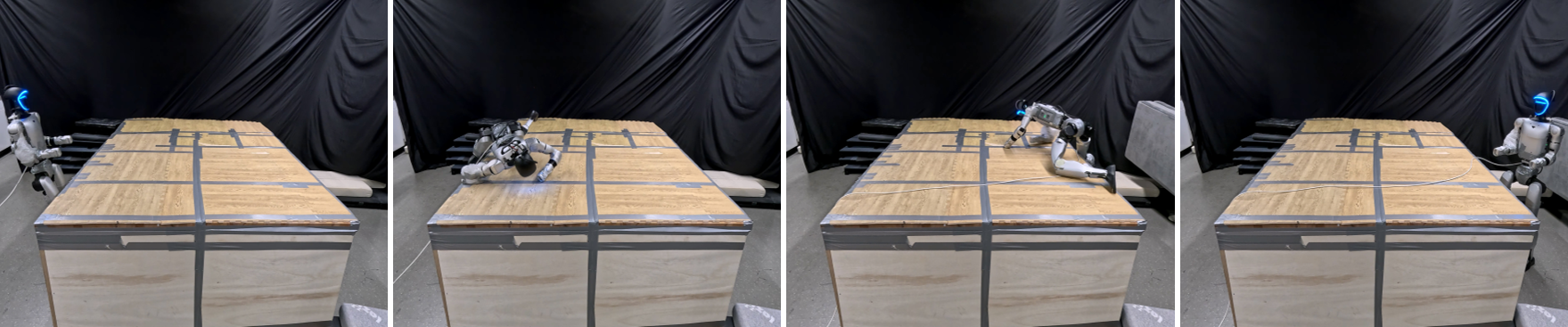}
        \caption{Robot completing platform traversal.}
        \label{fig:crawl_traversal}
    \end{subfigure}

    \caption{Experimental validation of the unified policy. The distilled system demonstrates real-world robustness by successfully completing diverse routes requiring complex full-body coordination.}
    \label{fig:traversal_group_fig}
\end{figure}

As shown in Fig. \ref{fig:traversal_group_fig}, we additionally demonstrate two new routes for high-platform traversal in the real world to further validate the context-aware capability of our proposed system. In both routes, the robot autonomously transitions between full-body maneuvers by perceiving the environmental geometry. For instance, when the robot is commanded to walk towards the platform, it perceives the obstacle and automatically triggers the climb-up skill to ascend. Similarly, when commanded to move towards the edge, the system perceives the drop and autonomously initiates the climb-down sequence to descend and reach a stable standing posture on the ground.

Route (a) includes the sequential execution of all six full-body maneuvers in the following order: walk (on the ground), climb-up, crawl, stand-up, walk (on the platform), lie-down, crawl, climb-down, and walk (on the ground). Our system achieves two full cycles of traversal consecutively, demonstrating the reliability of this context-aware gait switching.

Route (b) consists of a complete side-to-side traversal of the high platform with the following sequence: walk, climb-up, crawl, climb-down, and walk. Besides the successful execution of full-body skills, it also highlights the robustness of our perception pipeline in accurately identifying environmental contexts during dynamic maneuvers.

\subsection{Extended Experiments on Robustness}

The adaptability and robustness of our policy is showcased in three extreme cases:
(i) large external perturbation;
(ii) significant perception artifacts;
(iii) soft high platform;

\subsubsection{Robustness to Perception Artifacts}

\begin{figure}[h]
    \centering
    \includegraphics[width=1\linewidth]{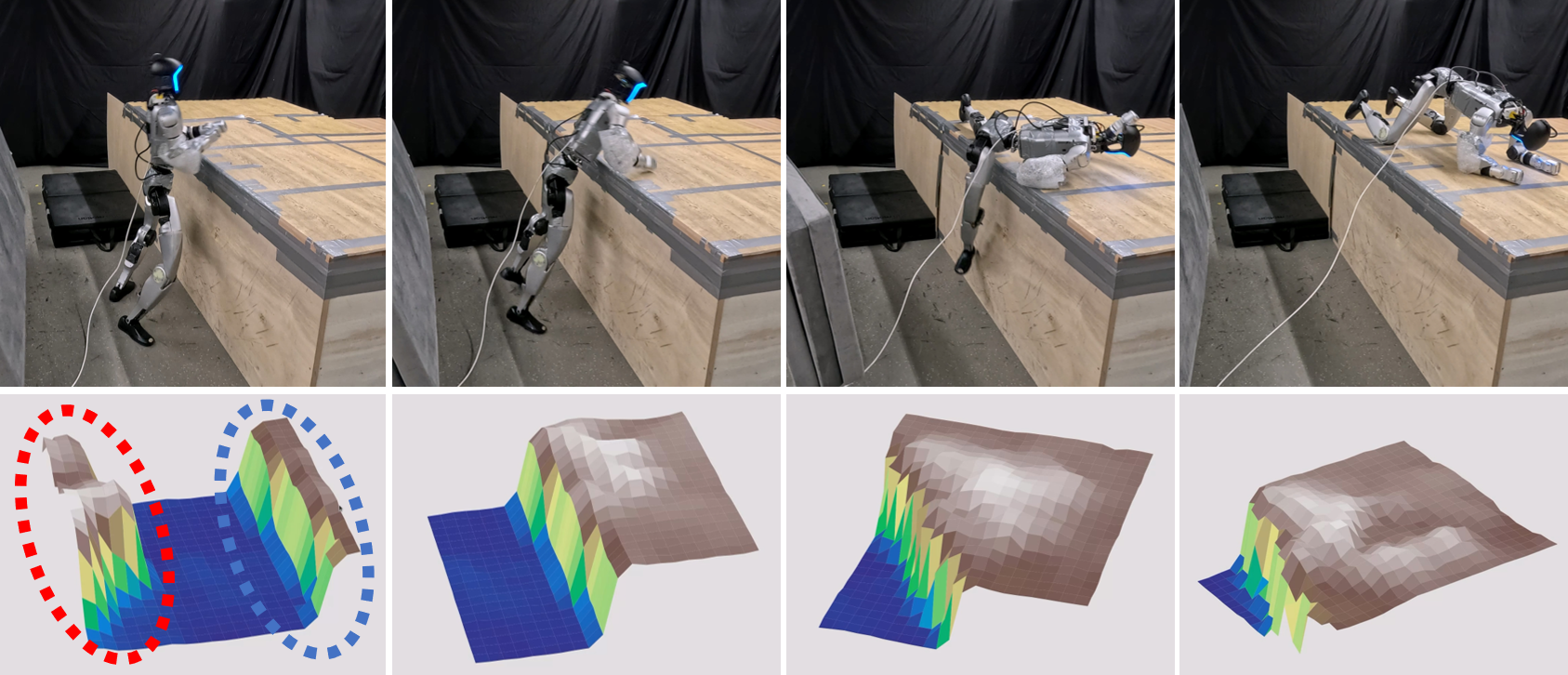}
    \caption{A typical artifacted elevation map with significant outlier cluster. The red circle denotes the outliers while the blue circle denotes the target platform. The map is aligned with the platform edge for a clearer view.}
    \label{fig:mapping_artifact}
\end{figure}

Fig. \ref{fig:mapping_artifact} illustrates a typical elevation map observed by the robot and the robot's corresponding climbing maneuver. The map contains a significant batch of ``ghost points'', which form a fake obstacle behind the robot of a scale comparable to the target platform. Despite these substantial perceptual artifacts, the robot successfully climbs up the platform with nominal movement, demonstrating the policy's perceptual robustness. This resilience is obtained from the noises injected during the training process, especially the outlier clusters, which showcases the necessity of rigorous perception and noise modeling for reliable real-world deployment. 

\begin{table}[h]
\centering
\caption{Comparative success rates of the climb-up skill on hardware. 
\textbf{SR}: success rate; \textbf{S/T}: successes/trials; 
\textbf{M.C.F.}: maximum contact force; 
\textbf{$H$ (m)}: platform height; 
\textbf{$A$ ($^\circ$)}: approach angle relative to the platform-edge normal.}
\label{tab:lidar_ablation}
\footnotesize
\setlength{\tabcolsep}{4pt}
\begin{tabular}{@{}l c c c c@{}}
\toprule
\textbf{Task} 
& \textbf{$H$ (m)} 
& \textbf{$A$ ($^\circ$)} 
& \textbf{S/T} 
& \textbf{SR (\%)} \\
\midrule
Climb-up (full system) 
& 0.8 
& $[-45, 45]$ 
& 15/15 
& 100 \\
\midrule
Climb-up (w/o drift \& outlier) \\
\quad (w/ post-processing) 
& 0.8 
& $[-45, 45]$ 
& 0/5 
& 0.0 \\
\midrule
Climb-up (w/ drift \& outlier) \\
\quad (w/o post-processing) 
& 0.8 
& $[-45, 45]$ 
& 3/5 
& 60.0 \\
\bottomrule
\end{tabular}
\end{table}

\begin{figure}[h]
\centering
  \includegraphics[width=0.5\linewidth]{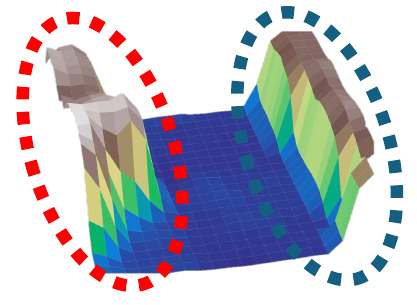}
  \caption{The red dashed region indicates severe mapping artifacts caused by outdated spatial measurements, while the blue dashed region denotes the target platform.
 }
  \label{fig:artifacted_map}
\end{figure}

To isolate the impact of the perception pipeline, we conduct an ablation study by selectively removing individual mapping pipeline components. As reported in Tab.~\ref{tab:lidar_ablation}, the full system achieves nearly $100\%$ success, whereas disabling real-time filtering and inpainting reduces the success rate to $60\%$. Moreover, a policy trained without simulated sensor drift and outlier corruption fails entirely on hardware. We observe that, in some scenarios, large regions of mapping artifacts remain in the elevation map (Fig.~\ref{fig:artifacted_map}), often caused by outdated spatial measurements near the robot that are not updated in time. Despite these severe artifacts, the robust policy still executes the intended maneuver correctly and successfully completes the task. These results demonstrate that modeling sensor artifacts during training is essential for tolerating mapping uncertainty and mitigating out-of-distribution (OOD) failures by expanding the effective training distribution. In addition, real-time map reconstruction is critical to prevent the policy from encountering severely corrupted inputs, such as outlier clusters or NaN holes, which can otherwise lead to catastrophic failures.

\subsubsection{Robustness to Varying Contact Property}

\begin{figure}[h]
    \centering
    \includegraphics[width=1\linewidth]{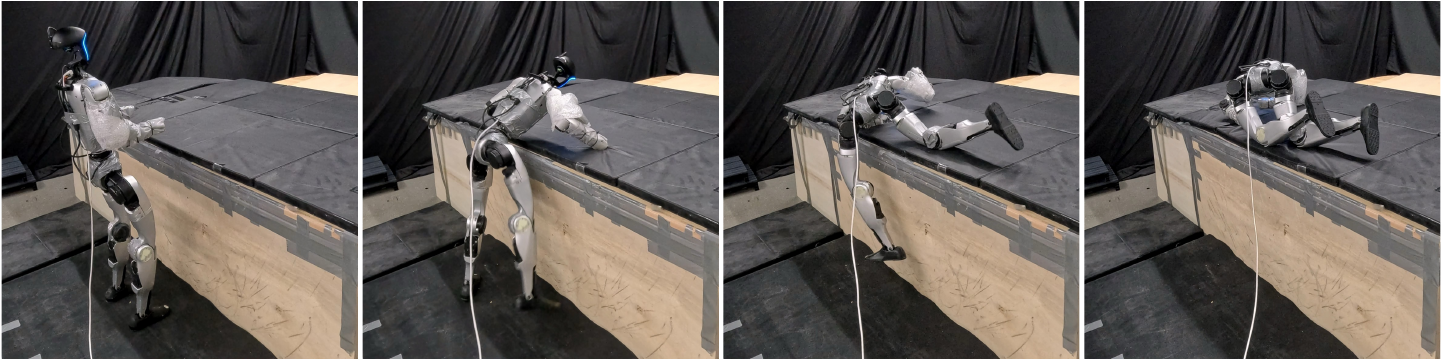}
    \caption{The robot successfully and stably climbs up the platform with a black soft mat on top.}
    \label{fig:softmat}
\end{figure}

We further evaluate the policy’s climbing capabilities by placing a soft mat, made of vinyl and foam, on top of the target platform (Fig. \ref{fig:softmat}). This setup challenges the robot to climb up an unseen material with significantly different compliance and friction properties compared to the rigid training environments. The robot successfully climbs the soft mat in the first trial, maintaining the same levels of stability and efficiency observed when climbing rigid platforms.

The stability of the climbing skill originates from the policy's quasi-static and contact-rich moving patterns. Unlike dynamic jumping or lunging behaviors, our policy does not rely on impulsive supporting forces provided by rigid surfaces, nor does it depend on a limited number of contact points to maintain equilibrium. Instead, the policy learns to distribute loads across multiple contact points, ensuring balance through consistent interaction with the platform. These results further demonstrate the adaptability of our policy to varying contact properties and highlight the advantages of our proposed reward formulation in learning safe, robust humanoid behaviors.


\subsection{Effectiveness of Multi-Teacher Distillation}

To evaluate our distillation pipeline, we compare the teacher and the distilled student across the four contact-rich full-body maneuvers in simulation. We report success rate and maximum contact force (Tab.~\ref{tab:teacher_student_compare}), evaluated over 1{,}000 parallel environments with the same randomization ranges used during training.
As expected, the specialized teacher policies achieve the strongest performance on their respective tasks. Nevertheless, the distilled student attains comparable success rates across all maneuvers, while maintaining statistically similar maximum contact forces within a safe range. Overall, these results indicate that the unified student policy faithfully captures the behaviors of diverse experts and retains near-teacher-level performance in simulation.

\begin{table}[h]
\centering
\caption{Comparison of Teacher and Student Performance. \textbf{SR:} Success Rate; \textbf{M.C.F.:} Max Contact Force. }
\label{tab:teacher_student_comp}
\footnotesize
\newcolumntype{C}{>{\centering\arraybackslash}X}
\begin{tabularx}{\columnwidth}{@{}l CCCC@{}}
\toprule
 & \multicolumn{2}{c}{\textbf{Teacher Policy}} & \multicolumn{2}{c}{\textbf{Student Policy}} \\
\cmidrule(lr){2-3} \cmidrule(lr){4-5}
\textbf{Skill} & \textbf{SR (\%)} & \textbf{M.C.F. (N)} & \textbf{SR (\%)} & \textbf{M.C.F. (N)} \\
\midrule
Climb-up   & 98.8 & $638 \pm 479$ & 98.6 & $657 \pm 324$ \\
Climb-down & 99.9 & $754 \pm 241$ & 99.0 & $762 \pm 539$ \\
Stand-up   & 99.5 & $632 \pm 222$ & 99.1 & $680 \pm 237$ \\
Lie-down   & 100.0 & $576 \pm 125$ & 100.0 & $637 \pm 124$ \\
\bottomrule
\end{tabularx}
\label{tab:teacher_student_compare}
\end{table}

\subsection{Details for Baseline Training}

We summarize the task reward formulations used by baseline methods in Tab.~\ref{tab:baseline_formulation}.

For the Velocity baseline, the commanded linear velocity satisfies
$v_{\mathrm{cmd}} \in [0.5, 1.0]$\,m/s in the world frame, and the goal position
$p_{\mathrm{goal}}$ is defined 0.5\,m inside the platform along the robot’s initial
heading. Distance (less Reg.) shares the same task reward as Distance but uses
contact-force regularization weights reduced by a factor of 10.

\begin{table}[t]
\centering
\caption{Formulations of the baseline task rewards.}
\label{tab:baseline_formulation}
\footnotesize
\begin{tabularx}{\columnwidth}{@{} l l @{}X c @{}}
\toprule
\textbf{Baseline} & \textbf{Reward} & \textbf{\quad Formulation} & \textbf{Weight} \\
\midrule
Velocity 
& Lin. Vel. Tracking 
& \quad$\exp\!\left(-\|v_{base,t}^{(xy)} - v_{cmd}^{(xy)}\|^2 / 0.5^2\right)$ 
& 4 \\
& Ang. Vel. Tracking 
& \quad$\exp\!\left(-(\omega_{base,t}^{(z)} - \omega_{cmd}^{(z)})^2 / 0.5^2\right)$ 
& 4 \\
\midrule
RND 
& Sparse Success 
& \quad$\mathbf{1}_{\{p_{LB,t}^{(z)} > h_{edge} \wedge p_{CoM,t}^{(x)} > x_{edge}\}}$ 
& 8 \\
\midrule
Distance 
& L2 Tracking 
& \quad$(1 + \|p_{base,t}^{(xy)} - p_{goal}^{(xy)}\|^2)^{-1}$ 
& 8 \\
\midrule
Direction 
& Cosine Sim. 
& $\quad\cos\!\left(\theta(v_{base,t},\, p_{goal} - p_{base,t})\right)$ 
& 8 \\
\midrule
Increment 
& Height Incr. 
& \quad$\mathbf{1}_{\{p_{LB,t}^{(z)} > p_{LB,t-1}^{(z)}\}} \mathbf{\cdot} \mathbf{1}_{\{x_t \notin x_g\}}$ 
& 4 \\
& Forward Incr. 
& \quad$\mathbf{1}_{\{p_{CoM,t}^{(x)} > p_{CoM,t-1}^{(x)}\}} \mathbf{\cdot} \mathbf{1}_{\{x_t \notin x_g\}}$ 
& 4 \\
\bottomrule
\end{tabularx}

\vspace{0.3em}

\vspace{-5pt}
\end{table}

\subsection{Details for Teacher Policy Training}

\subsubsection{Environment Configuration}

We train the teacher policy in IsaacLab using 4096 parallel environments for each single skill. While the six skills share basic observations, climbing skills are additionally based on height scan dots, and walking skill additionally utilize a phase signal to lead the gait pattern. The full list of observations is in Tab. \ref{tab:observation}. We also introduce perturbations and domain randomization, including previously discussed perception artifacts, to improve robustness (Tab. \ref{tab:perturbations}).

\begin{table}[h]
\centering
\caption{Observations and Noise for Teacher Training.}
\label{tab:observation}
\footnotesize
\begin{tabularx}{\columnwidth}{@{}l X l@{}}
\toprule
\textbf{Skill} & \textbf{Observation} & \textbf{Noise Range} \\
\midrule
All Skills     & Root angular velocity (rad/s) & $[-0.2, 0.2]$ \\
               & Projected gravity             & $[-0.05, 0.05]$ \\
               & Joint position (rad)          & $[-0.1, 0.1]$ \\
               & Joint velocity (rad/s)        & $[-1.5, 1.5]$ \\
               & Last action                   & --- \\
\midrule
Climb-up/down  & Elevation Map (m)             & refer to Tab. \ref{tab:perturbations} \\
\midrule
Walk / Crawl   & Phase signal                  & --- \\
               & Velocity Commands (m/s)       & --- \\
\bottomrule
\end{tabularx}
\end{table}

\begin{table}[h]
\centering
\caption{Perturbations and Domain Randomization Ranges.}
\label{tab:perturbations}
\footnotesize
\begin{tabularx}{\columnwidth}{@{}X l@{}}
\toprule
\textbf{Perturbed Terms} & \textbf{Perturbed Range} \\
\midrule
\multirow{2}{*}{Torso CoM Position (m)}     & $x, y: [-0.05, 0.05]$ \\
                                        & $z: [-0.02, 0.02]$ \\
Torso Mass (kg)                         & $m: [-1.0, 1.0]$ \\
\midrule
Static Friction                         & $\mu_s: [0.3, 1.6]$ \\
Dynamic Friction                        & $\mu_d: [0.3, 1.2]$ \\
Restitution                             & $e: [0.0, 0.5]$ \\
\midrule
Joint Default Position (rad)            & $q: [-0.01, 0.01]$ \\
Joint Initial Position (rad)            & $\dot q: [-0.15, 0.15]$ \\
\midrule
\multirow{5}{*}{External Push (m/s) (rad/s)}    & $v_x, v_y: [-0.5, 0.5]$ \\
                                        & $v_z: [-0.2, 0.2]$ \\
                                        & $\omega_r,\omega_p: [-0.5, 0.5]$ \\
                                        & $\omega_y: [-0.78, 0.78]$ \\
                                        & Interval (s): $[1, 3]$ \\
\midrule
\multirow{4}{*}{Elevation Map Noise (m)}   & $\text{Gaussian: } [-0.15, 0.15]$ \\
                                        & $\text{Drift } d_x,d_y: [-0.05, 0.05]$ \\
                                        & $\text{Drift } d_z: [-0.1, 0.05]$ \\
                                        & $\text{Outliers: } 20\%$ \\
\bottomrule
\end{tabularx}
\end{table}

The six skills are categorized into two groups:
(i) Non-periodic full-body maneuvers: climb-up, climb-down, stand-up, lie-down; 
(ii) Periodic locomotion skills: walk, crawl; 
Each group widely shares common rewards with several task-related reward terms. The full list of rewards is defined in Tab. \ref{tab:reward_group1}.

\subsubsection{Algorithm Design and Network Architecture}

We use Proximal Policy Optimization (PPO) to optimize the actor and the critic during the teacher policy training stage. The network architecture and hyperparameter are listed in Tab. \ref{tab:hyperparameter}.

\begin{table}[h]
\centering
\caption{Hyperparameter of Teacher Policy}
\label{tab:hyperparameter}
\small
\setlength{\tabcolsep}{12pt} 
\renewcommand{\arraystretch}{1.1} 

\begin{tabular}{c c}
\toprule
\multicolumn{2}{c}{\textbf{Environment and Architecture}} \\
\midrule
Num. of Environments & 4096 \\
Episode Length & 350 / 1000 \\
Network Type & MLP \\
Activation & ELU \\
Actor Network & [512, 256, 128] \\
Critic Network & [512, 256, 128] \\
\midrule
\multicolumn{2}{c}{\textbf{PPO Optimization Parameters}} \\
\midrule
Num. Epochs & 5 \\
Num. Mini Batches & 4 \\
Num. Steps per Batch & 24 \\
Num. Steps per Env & 24 \\
Normalization & Observation \\
Learning Rate $lr$ & 1.0e-3 \\
Clip Parameter & 0.2 \\
Entropy Coefficient & 0.01 \\
Gamma $\gamma$ & 0.99 \\
Lambda $\lambda$ & 0.95 \\
Desired KL value & 0.01 \\
Max Gradient Norm & 1.0 \\
\bottomrule
\end{tabular}
\end{table}

\vspace{-5pt}


\subsection{Details for Multi-Teacher Distillation}

\begin{table}[h]
\centering
\small 
\caption{Environment Distribution for Distillation.}
\label{tab:distill_env}
\setlength{\tabcolsep}{1pt} 
\renewcommand{\arraystretch}{1.1} 

\begin{tabular}{cccc}
\toprule
\textbf{Skills} & \textbf{Env. Prop.} & \textbf{Terrains} & \textbf{Vel. Cmd. ($m/s$)} \\ \midrule
Walk    & 0.17                & Rough + Plane   & Omni.                 \\
Crawl   & 0.08                & Plane           & Omni.                 \\ 
\midrule
Stand-up + Walk     & 0.07      & Plane         & Zero + Omni.          \\
Walk + Climb-up     & 0.16      & Platform      & Forward               \\
Climb-up + Crawl    & 0.12      & Platform      & Forward + Lateral     \\ 
Crawl + Climb-down  & 0.20      & Platform      & Lateral               \\
Climb-down + Walk   & 0.15      & Platform      & Lateral + Backward    \\
Lie-down + Crawl    & 0.05      & Plane         & Zero + Omni.          \\ 
\bottomrule
\end{tabular}
\end{table}





\subsubsection{Algorithm Design and Network Architecture}

The network architecture and hyperparameter are in the Tab. \ref{tab:hyperparameter_distill}.
\begin{equation}
\label{eq: loss}
    \mathcal{L}(\theta) = \mathbb{E}_{o \sim \mathcal{D}} \left[ \left\| \pi_{\theta}(o) - \mathbf{a}_{\text{teacher}} \right\|_2^2 \right]
\end{equation}

\begin{table}[h]
\centering
\caption{Hyperparameter for Distillation}
\label{tab:hyperparameter_distill}
\small
\setlength{\tabcolsep}{12pt} 
\renewcommand{\arraystretch}{1.1} 

\begin{tabular}{c c}
\toprule
\multicolumn{2}{c}{\textbf{Environment and Architecture}} \\
\midrule
Num. of Environments & 1000 \\
Episode Length & 400 \\
Activation & ELU \\
Network Type & MLP \\
Student Network & [2048, 1024, 512, 256] \\
\midrule
\multicolumn{2}{c}{\textbf{Optimization Parameters}} \\
\midrule
BC Iterations & 4 \\
DAgger Iterations & 16 \\
Num. Epochs & 1500 \\
Num. Steps per Batch & 20000 \\
Num. Steps per Env & 400 \\
Normalization & Observation \\
Learning Rate $lr$ & 3.0e-4 \\
Action Noise Std & 0.1 \\
Gradient Length & 1.0 \\
Max Gradient Norm & 1.0 \\
\bottomrule
\end{tabular}
\end{table}

\begin{table*}[t]
\centering
\caption{Reward Formulations for Teacher Policy Training.}
\label{tab:reward_group1}
\small
\begin{tabularx}{\textwidth}{@{} l l X c @{}}
\toprule
\multicolumn{4}{c}{\textbf{Non-periodic full-body maneuvers}} \\
\midrule
\textbf{Skill} & \textbf{Reward} & \textbf{Formulation} & \textbf{Weight} \\
\midrule 
All Skills & Survival & $ \mathbf{1}_{\{\neg term\}}$ & 15 \\
& Termination & $ \mathbf{1}_{\{term, \neg timeout\}}$ & -800 \\
& Force Penalty & $ \exp(0.01 \cdot \max(0, \|F\| - 500)) - 1$ & -1 \\
& Head Safety & $ \exp(0.1 \cdot \|F_{head}\|) - 1$ & -1 \\
& Joint Limits & $ \|\max(0, |q_t| - q_{soft})\|_1$ & -10 \\
& Hip Deviation & $ \mathbf{1}_{\{|q_{hip,yaw}| > 1.5 \vee |q_{hip,roll}| > 1.4\}}$ & -1 \\
& Waist Deviation & $ \mathbf{1}_{\{|q_{waist\_yaw}| > 1.4\}}$ & -6 \\
& Joint Velocity & $ \|\dot{q}_t\|^2$ & -0.001 \\
& Joint Accel. & $ \|\ddot{q}_t\|^2$ & -2e-8 \\
& Action Rate & $ \|a_t - a_{t-1}\|^2$ & -0.2 \\
& Torque & $ \|\tau_t\|^2$ & 1.5e-5 \\
& Power & $ \sum |\tau_t \cdot \dot{q}_t|$ & -1e-5 \\
& Body Slip & $ \sum_{i \in \mathcal{C}} \|v_{i,t}^{(xy)}\|$ & -0.1 \\
& Base Ang. Vel. & $ \|\omega_{base,t}^{(xy)}\|^2$ & -0.005 \\
& Base Accel. & $ \|\ddot{p}_{base,t}\|^2 + 0.02\|\dot{\omega}_{base,t}\|^2$ & -0.0001 \\
& Body Accel. & $ \sum_{i} \|\ddot{p}_{i,t}\|$ & -0.0002 \\
\midrule
Climb-up & Upward Progress & $\mathbf{1}_{\{p_{LB, max}^{(z)} \geq p_{LB,t}^{(z)}\}} \cdot \mathbf{1}_{\{x_{t} \notin x_{g}\}}$ & -4 \\
& Edge Approach & $\mathbf{1}_{\{p_{CoM,max}^{(x)} \geq p_{CoM,t}^{(x)}\}} \cdot \mathbf{1}_{\{x_{t} \notin x_{g}\}}$ & -4 \\
& Terminal Posture & $\mathbf{1}_{\{t>H-1s\}} \cdot \mathbf{1}_{\{x_{t} \in x_{g}\}} \cdot \exp(-0.1 \cdot \|q_t-q_{prone}\|)$ & 7 \\
\midrule
Climb-down & Descent Progress & $ \mathbf{1}_{\{p_{LB, min}^{(z)} \leq p_{LB, t}^{(z)}\}} \cdot \mathbf{1}_{\{x_t \notin x_g\}}$ & -4 \\
& Edge Clearance & $ \mathbf{1}_{\{p_{CoM, min}^{(x)} \leq p_{CoM, t}^{(x)}\}} \cdot \mathbf{1}_{\{x_t \notin x_g\}}$ & -4 \\
& Terminal Posture & $\mathbf{1}_{\{t>H-1s\}} \cdot \mathbf{1}_{\{x_{t} \in x_{g}\}} \cdot \exp(-0.1 \cdot \|q_t-q_{standing}\|)$ & 7 \\
\midrule
Stand-up & Height Progress & $\mathbf{1}_{\{p_{head, max}^{(z)} \geq p_{head, t}^{(z)}\}} \cdot \mathbf{1}_{\{x_t \notin x_g\}}$ & -4 \\
& Balance Progress & $\mathbf{1}_{\{d_{bal, min} \leq d_{bal, t}\}} \cdot \mathbf{1}_{\{x_t \notin x_g\}}$ & -4 \\
& Terminal Posture & $\mathbf{1}_{\{t>H-1s\}} \cdot \mathbf{1}_{\{x_{t} \in x_{g}\}} \cdot \exp(-0.1 \cdot \|q_t-q_{standing}\|)$ & 7 \\
\midrule
Lie-down & Descent Progress & $\mathbf{1}_{\{p_{CoM, min}^{(z)} \leq p_{CoM, t}^{(z)}\}} \cdot \mathbf{1}_{\{x_t \notin x_g\}}$ & -4 \\
& Head Placement & $\mathbf{1}_{\{p_{head, min}^{(z)} \leq p_{head, t}^{(z)}\}} \cdot \mathbf{1}_{\{x_t \notin x_g\}}$ & -4 \\
& Terminal Posture & $\mathbf{1}_{\{t>H-1s\}} \cdot \mathbf{1}_{\{x_{t} \in x_{g}\}} \cdot \exp(-0.1 \cdot \|q_t-q_{prone}\|)$ & 7 \\

\toprule
\multicolumn{4}{c}{\textbf{Periodic locomotion skills}} \\
\midrule
\textbf{Skill} & \textbf{Reward} & \textbf{Formulation} & \textbf{Weight} \\
\midrule
All Skills & Track lin. Velocity & $\exp(-\|v^{(x,y)} - v_{cmd}\|^2 / 0.5^2)$ & 1.3 \\
& Track Ang. Velocity & $\exp(-\|\omega^{(z)} - \omega_{cmd}\|^2 / 1.0^2)$ & 1.3 \\
& Vertical Lin. Velocity & $\|v^{(z)}\|^2$ & -2 \\
& Horizontal Ang. Velocity & $\|\omega^{(x,y)}\|^2$ & -0.15 / -0.05 \\
& Height Penalty & $(p_{root}^{(z)} - h_{des})^2$ & -10 \\
& Joint Acc. Penalty & $\sum_{j \in \mathcal{A}} \ddot{q}_{j,t}^2$ & $-2.5e-7$ \\
& Joint Vel. Penalty & $\sum_{j \in \mathcal{A}} \dot{q}_{j,t}^2$ & $-1.5e-3$ \\
& Action Rate & $\|a_t - a_{t-1}\|^2$ & -0.1 \\
& Joint Limits & $\sum (\max(0, q_{min} - q) + \max(0, q - q_{max}))$ & -5 \\
& Survival & $\mathbf{1}_{\{\neg terminated\}}$ & 0.2 / 10 \\
& Torque Penalty & $\|\tau_t\|^2$ & $-1.0e-5$ \\
& Undesired Contact & $\sum_{b \in \mathcal{B}} \mathbf{1}_{\{\|F_{contact, b}\| > 0.1\}}$ & -1 \\

\midrule
Walk & Base Orientation & $\|g_{b}^{(x,y)}\|^2$ & -1.0 \\
& Hip Deviation & $\|q_t - q_{default}\|^2$ & -1.0 \\
& Contact Slip & $\sum_{b \in \text{ankle\_roll}} \mathbf{1}_{\{\|F_{c,b}\| > 1\}} \cdot \|v_b\|^2$ & -0.2 \\
& Feet Swing Height & $\sum_{f \in \text{ankle\_roll}} \mathbf{1}_{swing} \cdot (0.08 - h_{f})^2$ & -20 \\
& Gait Phase & $\sum_{i} \mathbf{1}_{\{C_i = C_{target}(\phi_t)\}}$ & 0.18 \\
& Feet Air Time & $\sum_{f \in \text{ankle\_roll}} (\tau_{air,f} - 0.5) \cdot \mathbf{1}_{contact} \cdot \mathbf{1}_{move}$ & 0.1 \\

\midrule
Crawl & Termination & $\mathbf{1}_{terminated}$ & -100 \\
& Lying Deviation & $\|q_t - q_{lying}\|^2$ & -1.0 \\
& Contact Force Penalty & $\sum_{b} \max(0, \|F_{b}\| - 500)$ & -0.01 \\

\bottomrule
\end{tabularx}
\end{table*}

\begin{figure}[t]
    \centering
    \includegraphics[width=\linewidth]{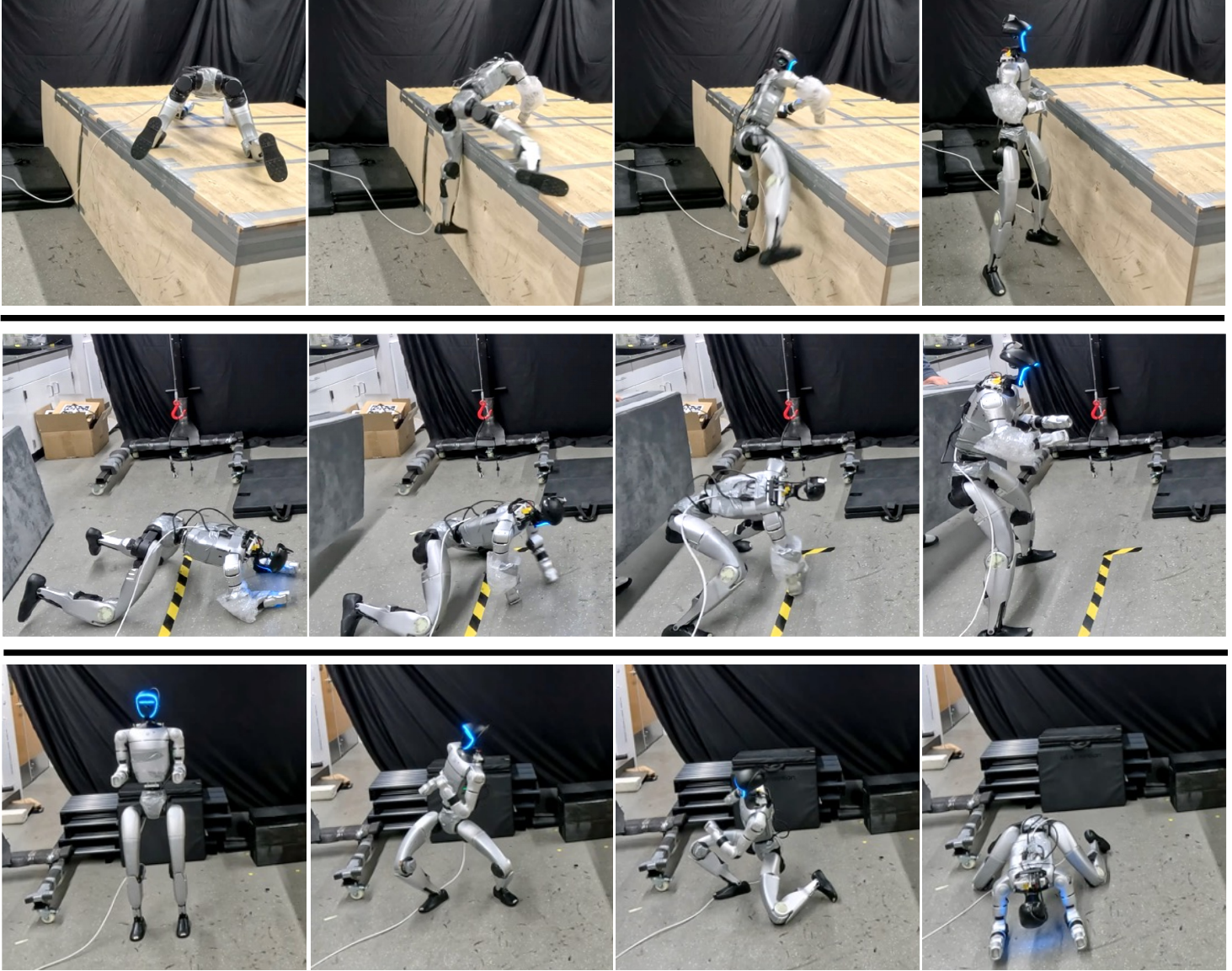}
    \caption{Three full-body maneuvers deployed on hardware: climb-down, stand-up, and lie-down.} 
    \label{fig:real_skills}
\end{figure}

We conduct all experiments with a 29-DoF Unitree G1 humanoid robot in both simulation and on hardware. Simulation environments are implemented in NVIDIA Isaac Sim. For real-world deployment, terrain perception is handled by an Intel Core i7 CPU, which processes data from a Livox MID-360 LiDAR to generate real-time elevation maps.

\subsection{Demonstration of Individual Skills}
Beyond climb-up, Fig.~\ref{fig:real_skills} presents the motion sequences of the remaining three full-body maneuvers.
\end{document}